\documentclass[journal]{IEEEtran}
\ifCLASSINFOpdf
\else
\fi
\usepackage{amsmath}
\usepackage{amsfonts}

\usepackage{float}

\usepackage{graphicx}
\usepackage{svg}

 \usepackage{booktabs} 
 \usepackage [utf8]{inputenc}

\begin{document}
\title{Vision Language Models in Medicine}

\author{Béria~Chingnabé~Kalpélbé, 
        Angel~Gabriel~Adaambiik,
        Wei~Peng% <-this % stops a space
% \thanks{M. Shell was with the Department
% of Electrical and Computer Engineering, Georgia Institute of Technology, Atlanta,
% GA, 30332 USA e-mail: (see http://www.michaelshell.org/contact.html).}% <-this % stops a space
% \thanks{J. Doe and J. Doe are with Anonymous University.}% <-this % stops a space
% \thanks{Manuscript received April 19, 2005; revised August 26, 2015.}
}


\markboth{Vision Language Models in Medicine}%
{Shell \MakeLowercase{\textit{et al.}}: Bare Demo of IEEEtran.cls for IEEE Journals}



\maketitle

\begin{abstract}
With the advent of Vision-Language Models (VLMs), medical artificial intelligence (AI) has experienced significant technological progress and paradigm shifts. This survey provides an extensive review of recent advancements in Medical Vision-Language Models (Med-VLMs), which integrate visual and textual data to enhance healthcare outcomes. We discuss the foundational technology behind Med-VLMs, illustrating how general models are adapted for complex medical tasks, and examine their applications in healthcare. The transformative impact of Med-VLMs on clinical practice, education, and patient care is highlighted, alongside challenges such as data scarcity, narrow task generalization, interpretability issues, and ethical concerns like fairness, accountability, and privacy. These limitations are exacerbated by uneven dataset distribution, computational demands, and regulatory hurdles. Rigorous evaluation methods and robust regulatory frameworks are essential for safe integration into healthcare workflows. Future directions include leveraging large-scale, diverse datasets, improving cross-modal generalization, and enhancing interpretability. Innovations like federated learning, lightweight architectures, and Electronic Health Record (EHR) integration are explored as pathways to democratize access and improve clinical relevance. This review aims to provide a comprehensive understanding of Med-VLMs' strengths and limitations, fostering their ethical and balanced adoption in healthcare.
\end{abstract}


% \fi
\IEEEpeerreviewmaketitle

\section{Introduction}
% Hope this several question will help you with this paper:

% 1. What are Visual Language Models and what makes them different from the traditional ML methods?

% 2. Why extend LLM to VLM, How they did this? 

% 3. How are VLMs currently being applied in medical settings?

% 4. What are the strengths and limitations of VLMs in medicine?

% 5. What are the benchmarks? 

% 6. How do people solve the security issues for clinic applications? Accuracy and Reliability, 

% 7. How VLM can be integrated into Clinical Workflow?

% 8. How do we address potential biases in VLMs that could lead to disparities in patient care?

% 9. Future ?

\IEEEPARstart{L}everaging advanced algorithms and neural network architectures like Transformers~\cite{vaswani2023attentionneed}, AI has been empowered with strong reasoning ability and made tremendous progress in recent years. Breakthroughs in model design and training methodologies have allowed machines to excel in complex tasks, including Natural Language Processing (NLP) applications such as language translation, sentiment analysis, and text generation, achieving high accuracy and fostering intuitive human-computer interactions. Similarly, advancements in Computer Vision (CV) have empowered AI to analyze and interpret images, videos, and audio sequences with remarkable precision. In healthcare, Artificial Intelligence (AI) is revolutionizing medicine by enabling data-driven insights, improving diagnostics, and personalizing treatments~\cite{topol2019,Esteva2017,KOUROU20158}. These innovations have enabled significant applications, such as medical imaging analysis, disease diagnosis, pathology, radiology workflow optimization, and surgical assistance, transforming patient care and clinical workflows~\cite{empeek2024,pmc2021}.

The medical field faces unique challenges in data interpretation and decision-making for healthcare specialists; they must analyze diverse types of information including medical imaging (X-rays, MRIs, pathology slides), clinical notes, patient histories, and real-time observations. Medical images are critical for diagnostic checks and measurements, such as identifying anatomical abnormalities, quantifying disease progression, or assessing treatment efficacy. On the other hand, textual data, such as clinical notes, nurse evaluations, and patient histories, provide essential context for screening, understanding symptoms, and documenting disease progression. Textual outputs, such as radiology reports or discharge summaries, are equally vital, as they synthesize findings into actionable insights for clinicians. The complexity and volume of this multi-modal medical data often lead to cognitive overload, impacting the speed and accuracy of diagnoses. Traditional single-modality approaches, which treat images and text separately, fail to capture the intricate relationships between visual findings and clinical context. This limitation underscores the need for integrated vision-language models (VLMs) that can bridge the gap between these modalities\cite{bordes2024introductionvisionlanguagemodeling}, enabling more comprehensive and accurate decision-making in healthcare. This integrated approach promises to enhance clinical decision-making by providing more contextually informed insights and reducing the cognitive burden on healthcare providers.

\begin{figure*}[ht]
    \centering
    \includegraphics[width=\linewidth]{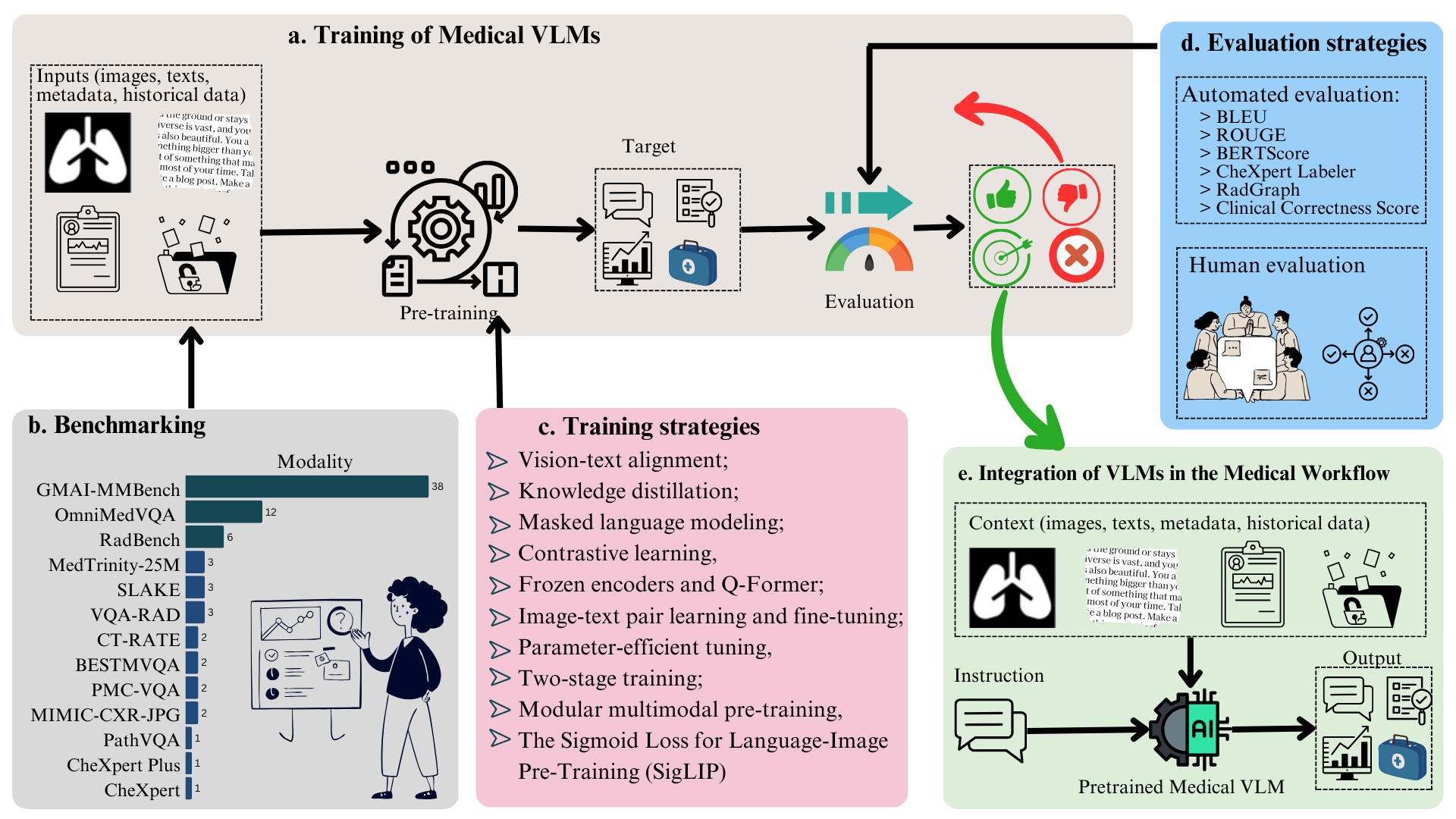}
    \caption{\textbf{Comprehensive Framework for Medical Vision-Language Models (VLMs)}. \textbf{(a)} Training involves processing diverse inputs such as images, texts, metadata, and historical data, followed by pre-training. \textbf{(b)} Benchmarking is conducted on a variety of medical datasets including GMAI-MMBench, OmniMedVQA, RadBench, and others. \textbf{(c)} Advanced training strategies are employed, such as vision-text alignment, knowledge distillation, masked language modeling, contrastive learning, and parameter-efficient tuning. \textbf{(d)} Evaluation strategies encompass automated metrics like BLEU, ROUGE, BERTScore, and clinical-specific tools like CheXpert Labeler and RadGraph, alongside human evaluation. \textbf{(e)} Integration of VLMs into the medical workflow leverages contextual data to provide actionable insights and improve clinical decision-making.}
    \label{fig:method}
\end{figure*}

However, visual and language provide totally different modalities that are not trivial to be integrated directly. As illustrated in Fig.~\ref{fig:method}, existing works address this challenge through various strategies, including vision-text alignment (in MedViL\cite{devlin2019bertpretrainingdeepbidirectional}, MedCLIP\cite{radford2021learningtransferablevisualmodels}, BioMedCLIP\cite{zhang2024biomedclipmultimodalbiomedicalfoundation}, VividMed\cite{luo2024vividmedvisionlanguagemodel}), knowledge distillation with VividMed\cite{luo2024vividmedvisionlanguagemodel}, masked language modeling (in MedViL\cite{devlin2019bertpretrainingdeepbidirectional} and BioMedCLIP\cite{zhang2024biomedclipmultimodalbiomedicalfoundation}) and contrastive learning in MedCLIP\cite{radford2021learningtransferablevisualmodels}, BioViL\cite{Boecking2022} \& ConVIRT\cite{zhang2022contrastivelearningmedicalvisual}. More recent advancements have introduced additional approaches, such as frozen encoders and Q-Former (e.g., BLIP-2\cite{li2023blip2bootstrappinglanguageimagepretraining}, InstructBLIP\cite{dai2023instructblipgeneralpurposevisionlanguagemodels}), image-text pair learning and fine-tuning (e.g., LLaVA\cite{liu2023visualinstructiontuning}, LLaVA-Med\cite{li2023llavamedtraininglargelanguageandvision}, BiomedGPT\cite{Zhang2024}, MedVInT\cite{zhang2024pmcvqavisualinstructiontuning}), parameter-efficient tuning (e.g., LLaMA-Adapter-V2\cite{zhang2024llamaadapterefficientfinetuninglanguage}), two-stage training (e.g., MiniGPT-4\cite{zhu2023minigpt4enhancingvisionlanguageunderstanding}), and modular multimodal pre-training (e.g., mPLUG-Owl\cite{ye2024mplugowlmodularizationempowerslarge}, Otter\cite{li2023ottermultimodalmodelincontext}) and the Sigmoid Loss for Language-Image Pre-Training (SigLIP)\cite{zhai2023sigmoidlosslanguageimage}. SigLIP replaces traditional softmax-based contrastive learning with a simpler sigmoid loss approach, enabling more efficient and scalable training by treating image-text pair alignment as a binary classification task. These methods aim to establish coherent relationships between visual inputs and textual outputs, enabling models to effectively interpret and generate relevant information across modalities. As a comparison, the contrastive learning-based methods leverage the similarities and differences between paired visual and textual data to enhance model robustness and generalization.

This paper provides a comprehensive review of VLMs and their applications in healthcare. We first discuss how VLMs are constructed by integrating advancements in NLP and computer vision. Next, we summarize key methodologies and advancements in the field, including state-of-the-art models like Qwen-VL\cite{bai2023qwenvlversatilevisionlanguagemodel}, RadFM\cite{wu2023generalistfoundationmodelradiology}, and DeepSeek-VL\cite{lu2024deepseekvlrealworldvisionlanguageunderstanding}. We then explore how VLMs are applied in the medical domain, highlighting their potential to improve diagnostic accuracy, clinical decision-making, and other healthcare tasks. Finally, we conclude by outlining future directions and challenges in the integration of VLMs into healthcare practices.

\section{Multi-modal Visual Text Models}
Vision-Language Models (VLMs) are a class of artificial intelligence models designed to process and integrate both medical imaging data (e.g., radiographs, histopathological slides) and clinical text (e.g., diagnostic reports, physician notes). The evolution of modern VLMs is rooted in advances in natural language processing, particularly the development of Bidirectional Encoder Representations of Transformers (BERT) \cite{devlin2019bertpretrainingdeepbidirectional}. BERT was originally developed for text classification and other language-specific tasks, including clinical text processing and medical literature analysis. Researchers soon recognized its potential for multimodal applications, enabling the integration of medical imaging with textual data within a unified framework. This advancement has paved the way for AI models capable of enhancing diagnostic interpretation, clinical decision support, and medical research.

\begin{figure}[ht]
    \centering
    \includegraphics[width=\linewidth]{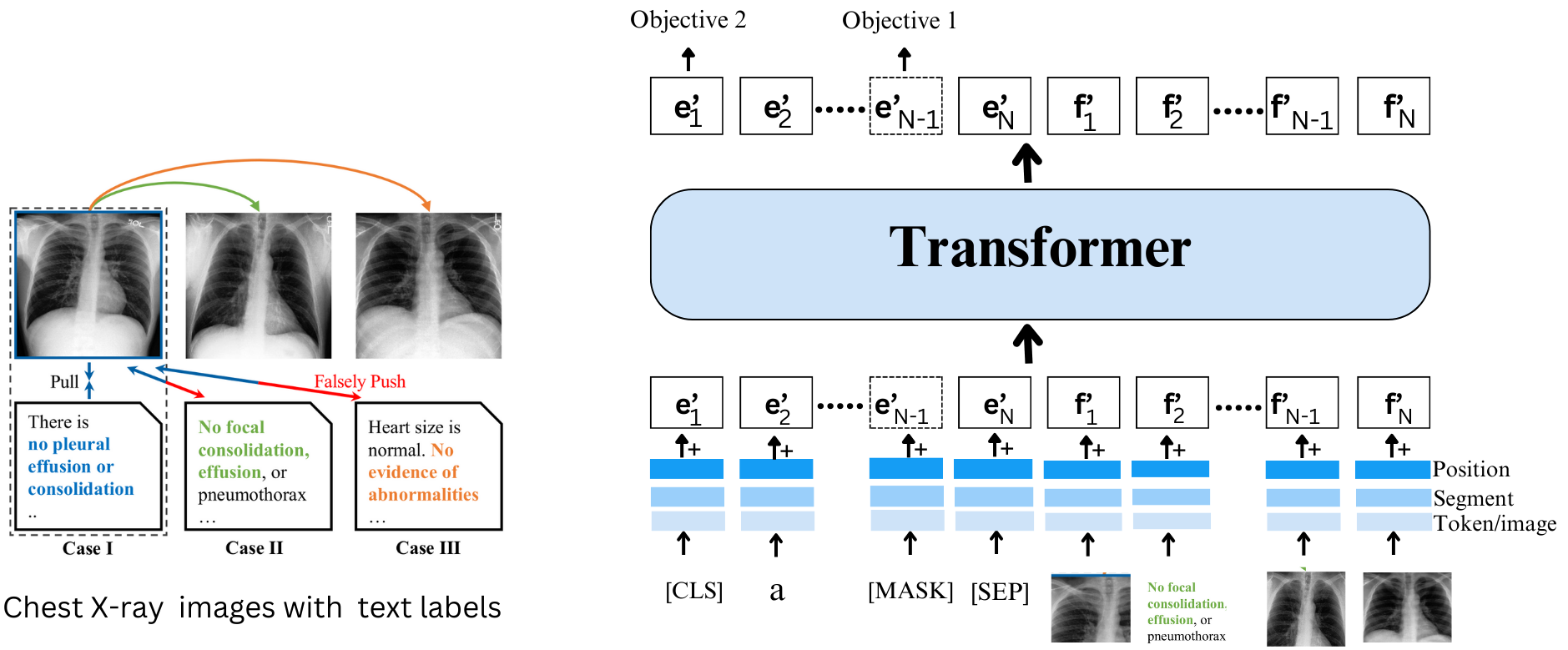}
    \caption{\textbf{Architecture of VisualBERT\cite{li2019visualbertsimpleperformantbaseline}.} This model integrates visual and textual inputs using a transformer-based architecture. Text tokens (e.g., "No focal consolidation, effusion or pneumothorax") and visual features extracted from the corresponding image are combined, along with positional and segment embeddings. The model is trained with dual objectives: masked language modeling (Objective 1) and visual-text alignment (Objective 2). This allows VisualBERT to effectively learn contextual representations that align both modalities for downstream tasks.}
    \label{fig:visualBERTarchitecture}
\end{figure}

\subsubsection{\textbf{Expanding BERT to Visual Data:} Several models extended BERT’s text-processing capabilities to incorporate visual data, enabling applications in medical imaging and clinical decision support. A notable early example is VisualBERT  \cite{li2019visualbertsimpleperformantbaseline}, which applies BERT’s transformer architecture to jointly model vision and language tasks. VisualBERT operates by stacking transformer layers, taking both a medical image (e.g., radiographs or histopathology slides) and its corresponding text (such as diagnostic reports or clinical notes) as input, and applying self-attention to learn interactions between the two modalities. . As illustrated in Figure \ref{fig:visualBERTarchitecture}, VisualBERT effectively captures the semantic alignment between visual and textual representations, making it a valuable tool for tasks such as automated medical report generation, radiology interpretation, and clinical decision-making support. }

After VisualBERT, numerous other models were developed, each bringing different innovations to vision-language modeling.

ViLBERT (Vision-and-Language BERT) \cite{lu2019vilbertpretrainingtaskagnosticvisiolinguistic} introduces two parallel processing streams for medical images and clinical text. These streams interact through a co-attention mechanism, allowing for more refined cross-modal representations. ViLBERT is particularly effective in tasks such as  visual question answering (VQA), where integrating imaging findings with textual clinical context is essential for diagnostic reasoning.

LXMERT \cite{tan2019lxmertlearningcrossmodalityencoder} 
further refines this approach by explicitly modeling anatomical and pathological relationships within medical images, using region-based visual features to enhance the understanding of complex medical scenes in conjunction with clinical text. It employs separate transformer networks for medical imaging and textual data before aligning the two streams, enabling more precise integration of medical findings with diagnostic reports. 

UNITER \cite{chen2020uniteruniversalimagetextrepresentation} adopts a unified approach, learning a joint embedding space for both medical images and clinical text without relying on separate transformers for each modality. This unified representation enhances cross-modal retrieval and multimodal reasoning, making it particularly useful for tasks such as automated radiology report generation, pathology image interpretation, and clinical decision support. 

Pixel-BERT \cite{huang2020pixelbertaligningimagepixels} 
deviates from using object-level visual features by directly processing raw imaging data at the pixel level, aligning pixel-based features with text representations. This approach eliminates the need for object detection preprocessing steps, allowing for a more generalizable model applicable to various imaging modalities, such as radiology, pathology, and ophthalmology 

% \subsubsection{Video and Multimodal Extensions}    
While the previously mentioned models focus on static medical images, some frameworks extend to video-based medical data. 
VideoBERT \cite{sun2019videobertjointmodelvideo} incorporates temporal and visual aspects of medical videos, such as ultrasound sequences, and endoscopic footage, aligning them with corresponding textual data like procedural notes or transcriptions. Similarly,  VD-BERT \cite{wang2020vdbertunifiedvisiondialog} extends BERT-like architectures to video understanding by integrating frame-level visual features with associated clinical text, enabling applications such as automated surgical video analysis, real-time diagnostic assistance, and video-based medical question answering.

\subsubsection{Specialized VLMs for Specific Domains}While some vision-language models are designed for specific industries, such as 
Fashion-BERT \cite{gao2020fashionberttextimagematching} for fashion-related tasks, specialized VLMs have also emerged in the medical domain. These models integrate visual data from medical imaging with textual descriptions, enhancing applications such as automated diagnosis, clinical decision support, and medical report generation. Similarly, M-BERT (Multimodal BERT) expands the applicability of vision-language integration across multiple medical modalities, improving AI-driven healthcare solutions. Each of these advancements represents incremental progress in bridging the gap between visual and textual understanding, extending the impact of VLMs across diverse fields, including medicine. In the healthcare sector, VLMs have demonstrated significant potential in various applications, such as automatically generating descriptive captions for medical images (e.g., X-rays, MRIs, and CT scans), assisting with diagnosis by interpreting imaging data and providing AI-driven diagnostic suggestions, generating coherent medical reports, and enabling cross-modal retrieval by linking medical images with corresponding clinical notes or retrieving relevant imaging studies based on textual queries.

\section{Core Concepts of Visual Language Modeling}
VLMs represent a transformative advancement in AI, bridging the gap between visual and textual data to enable multimodal understanding and reasoning. These models leverage the synergy between computer vision and NLP, allowing them to interpret, analyze, and generate insights from complex datasets that combine images and text. In the medical domain, VLMs have emerged as powerful tools for tasks such as medical image analysis, report generation, and visual question answering (VQA). By integrating visual and textual modalities, VLMs can enhance diagnostic accuracy, streamline clinical workflows, and support medical education. This section explores the core concepts underlying VLMs, their state-of-the-art implementations, and their diverse applications in healthcare as summarized in the table \ref{tab:vlm_medicine}.

\begin{table*}[ht]
    % \centering
    \caption{State of the art VLMs and their applications in medicine}
    \label{tab:vlm_medicine}
    \begin{tabular}{p{0.2\textwidth}p{0.25\textwidth}p{0.2\textwidth}p{0.25\textwidth}}
        \toprule
        \textbf{Baseline} & \textbf{Training Approach} & \textbf{Key Benchmarks} & \textbf{Primary Applications} \\
        \midrule
        BLIP-2 \cite{li2023blip2bootstrappinglanguageimagepretraining}, InstructBLIP\cite{dai2023instructblipgeneralpurposevisionlanguagemodels} & Frozen encoders, Q-Former & COCO\cite{lin2015microsoftcococommonobjects}, VQA-Med\cite{Ben2019} & Medical Image Captioning, VQA \\
        \midrule
        LLaVA\cite{liu2023visualinstructiontuning}, LLaVA-Med\cite{li2023llavamedtraininglargelanguageandvision}, BiomedGPT\cite{Zhang2024}, MedVInT\cite{zhang2024pmcvqavisualinstructiontuning} & Image-text pair learning, fine-tuning & VQA-RAD\cite{Lau2018}, SLAKE\cite{liu2021slakesemanticallylabeledknowledgeenhanceddataset} & VQA, Clinical Reasoning \\
        \midrule
        LLaMA-Adapter-V2\cite{zhang2024llamaadapterefficientfinetuninglanguage} & Parameter-efficient tuning & Multi-modal datasets & Multimodal Instruction Following \\
        \midrule
        MiniGPT-4\cite{zhu2023minigpt4enhancingvisionlanguageunderstanding} & Two-stage training & Vicuna\cite{zhu2023minigpt4enhancingvisionlanguageunderstanding} & Advanced Image Understanding \\
        \midrule
        mPLUG-Owl\cite{ye2024mplugowlmodularizationempowerslarge}, Otter\cite{li2023ottermultimodalmodelincontext} & Multimodal pre-training & MIMIC-IT\cite{li2023mimicitmultimodalincontextinstruction} & VQA, Medical Dialogue \\
        \midrule
        Qwen-VL\cite{bai2023qwenvlversatilevisionlanguagemodel} & Three-stage training & MIMIC-CXR\cite{Johnson2019}, CheXpert\cite{irvin2019} & Medical Image Captioning, VQA \\
        \midrule
        MedViLL\cite{devlin2019bertpretrainingdeepbidirectional}, MedCLIP\cite{radford2021learningtransferablevisualmodels}, BioViL\cite{Boecking2022}, BioMedCLIP\cite{zhang2024biomedclipmultimodalbiomedicalfoundation}, ConVIRT\cite{zhang2022contrastivelearningmedicalvisual}, VividMed\cite{luo2024vividmedvisionlanguagemodel}, Flamingo-CXR\cite{Tanno2024}, Med-Flamingo\cite{moor2023medflamingomultimodalmedicalfewshot} & Contrastive learning, Medical image-text pair learning, Unsupervised pre-training, Three-stage training, Fine-tuning & MIMIC-CXR\cite{Johnson2019}, MS-CXR\cite{Boecking2022}, ChestX-ray\cite{Wang_2017}, VinDr-CXR\cite{nguyen2022vindrcxropendatasetchest}, IND1\cite{Tanno2024} & Medical Report Generation, VQA, Radiology Task Performance, Medical Image Classification, Retrieval, Image Captioning \\
        \midrule
        RadFM\cite{wu2023generalistfoundationmodelradiology} & Pre-trained on MedMD & RadBench\cite{wu2023generalistfoundationmodelradiology} & Medical Diagnosis, VQA \\
        \midrule
        DeepSeek-VL\cite{lu2024deepseekvlrealworldvisionlanguageunderstanding} & Three-stage training: adaptor, joint pretraining, fine-tuning & Common Crawl,Web Code, E-books, arXiv Articles, ScienceQA, ScreenQA, etc.   &  Multimodal understanding and reasoning\\
        \bottomrule
    \end{tabular}
\end{table*}

\subsection{Sate-of-the-art VL models}

Recent advancements in VLMs have led to groundbreaking architectures like BLIP-2, LLaVA, and MiniGPT-4. These models excel in tasks such as image captioning, visual question answering, and medical image analysis. This subsection examines the most influential VLMs, their methodologies, and their contributions to multimodal AI.

\subsubsection{\textbf{BLIP-2}}
 Bootstrapping Language-Image Pre-training (BLIP-2) \cite{li2023blip2bootstrappinglanguageimagepretraining} introduces a groundbreaking approach to vision-language pre-training that leverages frozen image encoders and LLMs to enhance performance across various vision-language tasks. This innovative method aims to achieve state-of-the-art results while minimizing trainable parameters, addressing the critical challenge of bridging visual and textual modalities. The research is fundamentally driven by the growing need for efficient multimodal AI systems that can effectively understand and generate language based on visual inputs, particularly in applications such as image captioning, visual question answering, and image-text retrieval.

The methodology of BLIP-2 centers on a two-stage pre-training process utilizing a Querying Transformer (Q-Former). The first stage emphasizes vision-language representation learning with a frozen image encoder, followed by a second stage focused on generative learning with a frozen LLM. This approach incorporates image-text contrastive learning to maximize mutual information between modalities, image-grounded text generation, and image-text matching to refine alignment between visual and textual elements. The methodology's efficiency is enhanced through the use of frozen models, which reduces computational costs and enables larger batch sizes during training. The implementation consistently employs the AdamW optimizer and cosine learning rate decay, with the overall design allowing for more samples per GPU compared to traditional end-to-end methods.

The experimental results reveal BLIP-2's superior performance across multiple benchmarks while maintaining a significantly reduced parameter count compared to existing models. The system demonstrates remarkable achievements in VQA tasks, outperforming models like Flamingo80B despite using 54 times fewer parameters. In image captioning and retrieval tasks, particularly on COCO datasets, BLIP-2 shows competitive performance and exhibits strong zero-shot capabilities in generating accurate text descriptions from unseen images. The model's architecture enables rapid pre-training, requiring less than a week of computational time on single GPU setups, while maintaining high efficiency and generalization across different tasks. The research also acknowledges potential limitations, including risks associated with frozen LLM outputs and the ongoing need for advancement in multimodal AI development.

\subsubsection{\textbf{LLaVa}}
The idea behind LLaVa\cite{liu2023visualinstructiontuning} is to develop an efficient foundation language model termed "LLaVA" (Language and Vision Assistant), which integrates both language understanding and visual comprehension. The primary objective is to enhance the capabilities of language models in multimodal contexts, enabling them to better interpret and generate text based on visual inputs. This involves not only improving performance in traditional language tasks but also expanding functionality to include complex visual reasoning and description tasks.

The authors employed a two-step training process. Initially, they pre-trained the LLaVA model on a filtered dataset comprising 595,000 image-text pairs, sourced from the CC3M dataset. This filtering process involved selecting noun-phrases based on their frequency to ensure diverse representation across concepts in the dataset. The model was pre-trained for one epoch using a learning rate of 2e-3 and a batch size of 128. Following pre-training, the model underwent fine-tuning on a specialized dataset, LLaVA-Instruct-158K, for three epochs with a reduced learning rate of 2e-5 and a batch size of 32. Various optimization techniques, such as using the Adam optimizer with no weight decay and enabling BF16 and TF32, were implemented to balance speed and precision during training.

The results demonstrate that the LLaVA model exhibits significant improvements in responding to multimodal tasks compared to previous models. It effectively generates detailed and contextually appropriate descriptions of images, showcasing an ability to understand and relate visual elements to textual instructions. Performance benchmarks indicate that LLaVA not only maintains high accuracy in traditional language tasks but also excels in complex visual reasoning scenarios, establishing it as a robust tool for applications requiring integration of language and vision.

\subsubsection{\textbf{LLaMa\_Adapter\_v2}}
LLaMA-Adapter\cite{zhang2024llamaadapterefficientfinetuninglanguage}, introduced in early 2023, was one of the pioneering attempts to extend Large Language Models (LLMs) to handle visual inputs through parameter-efficient adaptation. The model utilized a lightweight adapter architecture that could be trained while keeping the base LLaMA model frozen, making it computationally efficient. Its key innovation was the introduction of a perceiver-style architecture that processed visual inputs and injected them into the LLM's attention layers. However, the model faced limitations in complex visual reasoning tasks and struggled with detailed visual instruction following.

Building upon these foundations, LLaMA-Adapter-V2\cite{gao2023llamaadapterv2parameterefficientvisual} emerged as a parameter-efficient visual instruction model designed to enhance the capabilities of LLMs in handling multi-modal reasoning tasks. The goal is to improve the model's ability to follow visual instructions and perform complex visual reasoning, surpassing the limitations of previous models like LLaMA-Adapter and achieving performance closer to that of GPT-4.
The authors augmented the original LLaMA-Adapter by unlocking more learnable parameters and introducing an early fusion strategy to incorporate visual tokens into the early layers of the LLM. They also implemented a joint training paradigm using image-text pairs and instruction-following data, optimizing disjoint groups of learnable parameters. Additionally, they integrated expert models (e.g., captioning and OCR systems) to enhance image understanding without incurring additional training costs. It demonstrated superior performance in open-ended multi-modal instructions with only a small increase in parameters over the original LLaMA model. The new framework showed stronger language-only instruction-following capabilities and excelled in chat interactions. The model achieved significant improvements in multi-modal reasoning tasks, effectively balancing image-text alignment and instruction-following learning targets.

\subsubsection{\textbf{MiniGPT-4}}
MiniGPT-4\cite{zhu2023minigpt4enhancingvisionlanguageunderstanding}
is a vision-language model designed to replicate the advanced multi-modal capabilities of GPT-4. The authors aim to demonstrate that aligning visual features with an advanced LLM can enhance vision-language understanding and generation, similar to GPT-4’s abilities.

MiniGPT-4 aligns a frozen visual encoder with a frozen advanced LLM, Vicuna, using a single projection layer. The model is trained in two stages: initially on a large collection of image-text pairs to acquire vision-language knowledge, and then fine-tuned with a smaller, high-quality dataset to improve language generation reliability and usability.
It exhibits advanced capabilities such as generating detailed image descriptions, creating websites from hand-drawn drafts, and explaining visual phenomena. The model’s performance is significantly improved after the second-stage fine-tuning, demonstrating its potential to achieve vision-language tasks comparable to those of GPT-4.

\subsubsection{\textbf{mPLUG-Owl}}
A modularized multi-modal foundation model known as mPLUG-Owl\cite{ye2024mplugowlmodularizationempowerslarge}, is oriented aims to improve multimodal reasoning and action through advanced model architectures that allow for better comprehension and interaction across different data types. It focuses on tasks such as VQA and interaction in multi-round conversations.
The training procedure involves a two-stage scheme that includes multimodal pre-training and joint instruction tuning. This approach allows the model to develop a nuanced understanding of both visual and textual instructions, enhancing its performance on tasks that require integration of these modalities. The authors conducted quantitative and qualitative evaluations to assess the model's capabilities, comparing it against baseline models like MM-REACT and MiniGPT-4. A series of experiments focused on instruction understanding were performed, indicating that multimodal instruction tuning significantly improves the model's performance. The evaluation metrics included response accuracy and the ability to comprehend complex instructions involving spatial orientation and human behavior, which were systematically analyzed using datasets such as OwlEval.

mPLUG-Owl significantly outperformed baseline models in various multimodal tasks. For instance, in knowledge-intensive question answering, mPLUG-Owl was able to identify movie characters in images more accurately compared to other models. It showed a superior capability in multi-round conversations where it effectively responded to referential questions that required spatial and contextual reasoning. The quantitative analysis indicated that the model achieved the best performance metrics when both multimodal pre-training and joint instruction tuning were applied. Furthermore, the findings highlighted that while text-only instruction tuning improved comprehension, incorporating visual data was crucial for enhancing knowledge and reasoning capabilities.

\subsubsection{\textbf{Otter}}
Otter\cite{li2023ottermultimodalmodelincontext} is a multi-modal model based on OpenFlamingo\cite{awadalla2023openflamingoopensourceframeworktraining}, designed to improve instruction-following and in-context learning abilities. It is trained on the MIMIC-IT\cite{li2023ottermultimodalmodelincontext} dataset proposed in the same paper, which includes instruction-image-answer triplets and in-context examples. The training process involved fine-tuning specific layers while keeping the vision and language encoders frozen, optimizing the model to run on four RTX-3090 GPUs.

Otter demonstrated improved instruction-following and in-context learning capabilities compared to OpenFlamingo. It provided more detailed and accurate descriptions of images and better understood complex scenarios. The model’s performance was evaluated through various experiments, showing significant advancements in visual question answering and commonsense reasoning tasks.

\subsubsection{\textbf{InstructBLIP}}
InstructBLIP\cite{dai2023instructblipgeneralpurposevisionlanguagemodels} is a vision-language instruction tuning framework. A pre-trained BLIP-2 model and introduce an instruction-aware Query Transformer (Q-Former) are used to extract visual features tailored to given instructions. The authors gathered 26 publicly available datasets across 11 task categories, transforming them into instruction tuning format. The model is trained on 13 held-in datasets and evaluated on 13 held-out datasets to assess zero-shot performance. A balanced sampling strategy is employed to ensure synchronized learning progress across datasets.
It achieved state-of-the-art zero-shot performance on all 13 held-out datasets, significantly outperforming BLIP-2 and larger Flamingo models. The model also excels in fine-tuning on individual downstream tasks, such as achieving 90.7\% accuracy on ScienceQA questions with image contexts. Qualitative evaluations demonstrate InstructBLIP’s superior capabilities in complex visual reasoning, knowledge-grounded image description, and multi-turn visual conversations.

\subsubsection{\textbf{VPGTrans}}

a two-stage transfer framework designed to transfer a Visual Prompt Generator (VPG) across different Large Language Models (LLMs), VPGTrans\cite{zhang2023vpgtranstransfervisualprompt}. The first stage involves warming up the projector with a high learning rate to adapt the pre-trained VPG to a new LLM, preventing performance drops. The second stage is vanilla fine-tuning of both the VPG and projector. This approach aims to reduce computational costs and training data requirements compared to training a VPG from scratch.

The VPGTrans framework demonstrated significant efficiency improvements, achieving up to 10 times acceleration for small-to-large LLM transfers and up to 5 times acceleration for transfers between different model types. Notably, it enabled a BLIP-2 ViT-G OPT2.7B to 6.7B transfer with less than 10\% of GPU hours and 10.7\% of training data compared to original training. Additionally, VPGTrans outperformed the original models on several datasets, showing improvements in VQAv2 and OKVQA accuracy.

\subsubsection{\textbf{Qwen-VL}}
The Qwen-VL\cite{bai2023qwenvlversatilevisionlanguagemodel} series represents advanced large-scale vision-language models designed to understand and process both text and images. These models leverage a robust architecture that combines a large language model, a visual encoder, and a vision-language adapter, enabling them to perform a variety of tasks in the realm of vision and language.

Initially, the model undergoes pre-training using a large-scale dataset of image-text pairs, which is refined from an original set of five billion to 1.4 billion cleaned samples, predominantly in English and Chinese. In the second stage, multi-task pre-training is conducted, enhancing the input resolution of the visual encoder and employing interleaved image-text sequences. This phase utilizes a vast array of data for various tasks, including captioning and visual question answering, allowing the model to learn effectively from diverse sources. Finally, the supervised fine-tuning stage focuses on instruction tuning to improve the model's interaction capabilities, employing mixed dialogue data and manual annotations to enhance multi-image comprehension and localization skills. Throughout this process, the model architecture integrates a language-aligned visual encoder and a position-aware adapter, ensuring efficient processing of both visual and textual data, ultimately enabling the Qwen-VL-Chat model to perform a wide range of vision-language tasks with impressive accuracy and fine-grained understanding.

\subsubsection{\textbf{DeepSeek-VL}}
DeepSeek-VL\cite{lu2024deepseekvlrealworldvisionlanguageunderstanding} is an open-source vision-language model developed for real-world applications in vision and language understanding, available in two variants with 1.3B and 6.7B parameters. The model aims to enhance user experience in various scenarios by integrating diverse data and an efficient architecture.

The architecture is built on a diverse dataset, including web screenshots, PDFs, OCR content, and charts, to comprehensively represent real-world scenarios, with instruction-tuning datasets derived from real user interactions to enhance practical usability. The training pipeline involves three stages: initial training of the vision-language adaptor to link visual and textual elements, joint pretraining of the vision-language model, and supervised fine-tuning to refine multimodal capabilities. A modality warm-up strategy dynamically adjusts the ratio of visual and language data during training, preserving linguistic performance while improving multimodal understanding. The model employs a token economy, compressing high-resolution images into 576 tokens to balance visual richness and token efficiency, making it suitable for multi-turn inference. Designed for scalability, DeepSeek-VL plans to integrate Mixture of Experts (MoE) technology to further enhance its multimodal capabilities. It demonstrates exceptional performance across visually-centric benchmarks while maintaining strong language proficiency, surpassing existing generalist models. Additionally, its open-source availability encourages community-driven innovation and research.

DeepSeek-VL demonstrates state-of-the-art or competitive performance across various vision-language (VL) benchmarks at comparable model sizes, showcasing its robust multimodal capabilities. In language-centric tasks, it performs on par with or even surpasses its predecessor, DeepSeek-7B, achieving scores of 68.4 on HellaSwag and 52.4 on MMLU, underscoring its strong linguistic proficiency alongside visual understanding. However, the model exhibits a notable decline in mathematical reasoning tasks, scoring 18.0 on GSM8K compared to DeepSeek-7B's 55.0, highlighting a potential area for improvement. When compared to advanced models like GPT-4V, DeepSeek-VL-7B excels in areas such as Recognition, Conversion, and Commonsense Reasoning, but GPT-4V maintains an edge in logical reasoning tasks, indicating room for further refinement in complex reasoning capabilities. Overall, DeepSeek-VL establishes itself as a highly competitive model in the VL landscape, balancing strong performance across multiple domains while identifying specific areas for future enhancement.

\subsubsection{\textbf{Pre-training VL models with SigLIP}}
The Sigmoid Loss for Language-Image Pre-Training (SigLIP) \cite{zhai2023sigmoidlosslanguageimage} represents a significant shift in vision-language pre-training by replacing traditional softmax-based contrastive learning with a simpler and more efficient sigmoid loss approach. This innovation simplifies distributed loss computation and enhances training efficiency, making it particularly suitable for large-scale VL model training.

In traditional softmax-based contrastive learning, the objective function for training a VL model involves an image encoder $f$ and a text encoder $g$. The goal is to minimize the following loss:
\begin{equation}
    - \frac{1}{2 |\mathcal{B}|} \sum_{i=1}^{|\mathcal{B}|} 
\left( 
\underbrace{\log \frac{e^{t x_i \cdot y_i}}{\sum_{j=1}^{|\mathcal{B}|} e^{t x_i \cdot y_j}}}_{\text{image} \to \text{text softmax}}
+ 
\underbrace{\log \frac{e^{t x_i \cdot y_i}}{\sum_{j=1}^{|\mathcal{B}|} e^{t x_j \cdot y_i}}}_{\text{text} \to \text{image softmax}}
\right)
\end{equation}
Where $x_i=\frac{f(I_i)}{|| f(I_i) ||_2}$ and $y_i=\frac{g(T_i)}{|| g(T_i) ||_2}$ represent the normalized embeddings of the image $I_i$ and $T_i$ respectively. While effective, this approach requires computing pairwise similarities across the entire batch, which can be computationally expensive and complex to implement in distributed settings.

SigLIP addresses these limitations by reformulating the learning problem as a binary classification task. Instead of computing softmax probabilities over all pairs, SigLIP processes image-text pairs independently, assigning positive labels to matching pairs $(I_i, T_i)$ and negative labels to non-matching pairs $(I_i, T_{j\ne i})$. The objective function is then defined as:

\begin{equation}
    - \frac{1}{|\mathcal{B}|} \sum_{i=1}^{|\mathcal{B}|} \sum_{j=1}^{|\mathcal{B}|} 
\log \underbrace{\frac{1}{1 + e^{z_{ij} (-t x_i \cdot y_j + b)}}}_{\mathcal{L}_{ij}}
\end{equation}
Here, $z_{ij}$ is the label for a given image-text pair, where $z_{ij}=1$ for positive pairs and $z_{ij}=-1$ for negative pairs. The sigmoid loss directly optimizes the binary classification task, simplifying the training process and reducing computational overhead. This approach not only improves scalability but also maintains competitive performance in vision-language alignment tasks.

\subsection{Applications}
VLMs are particularly useful for healthcare applications that require the interpretation of visual data, such as medical images. They can also answer questions about this visual data, such as identifying anomalies in the medical data.

\subsubsection{\textbf{MedViLL}}
For the task of report generation from medical images, some models have been proposed in the literature. Medical Vision Language Learner (MedViLL) \cite{Moon_2022} for example, is a medical model based on BERT architecture combined with a novel multimodal attention masking scheme and maximizes the generalization of both vision-language understanding tasks including diagnosis classification, medical image-report retrieval, and medical visual question answering. It also enables vision-language generation task such as radiology report generation. MedViLL is pre-trained on MIMIC-CXR dataset \cite{Johnson2019}, containing 227,835 imaging studies for 65,379 patients presenting the Beth Israel Deaconess Medical Center Emergency Department between 2011–2016.

\begin{figure}[ht]
    \centering
    \includegraphics[width=\linewidth]{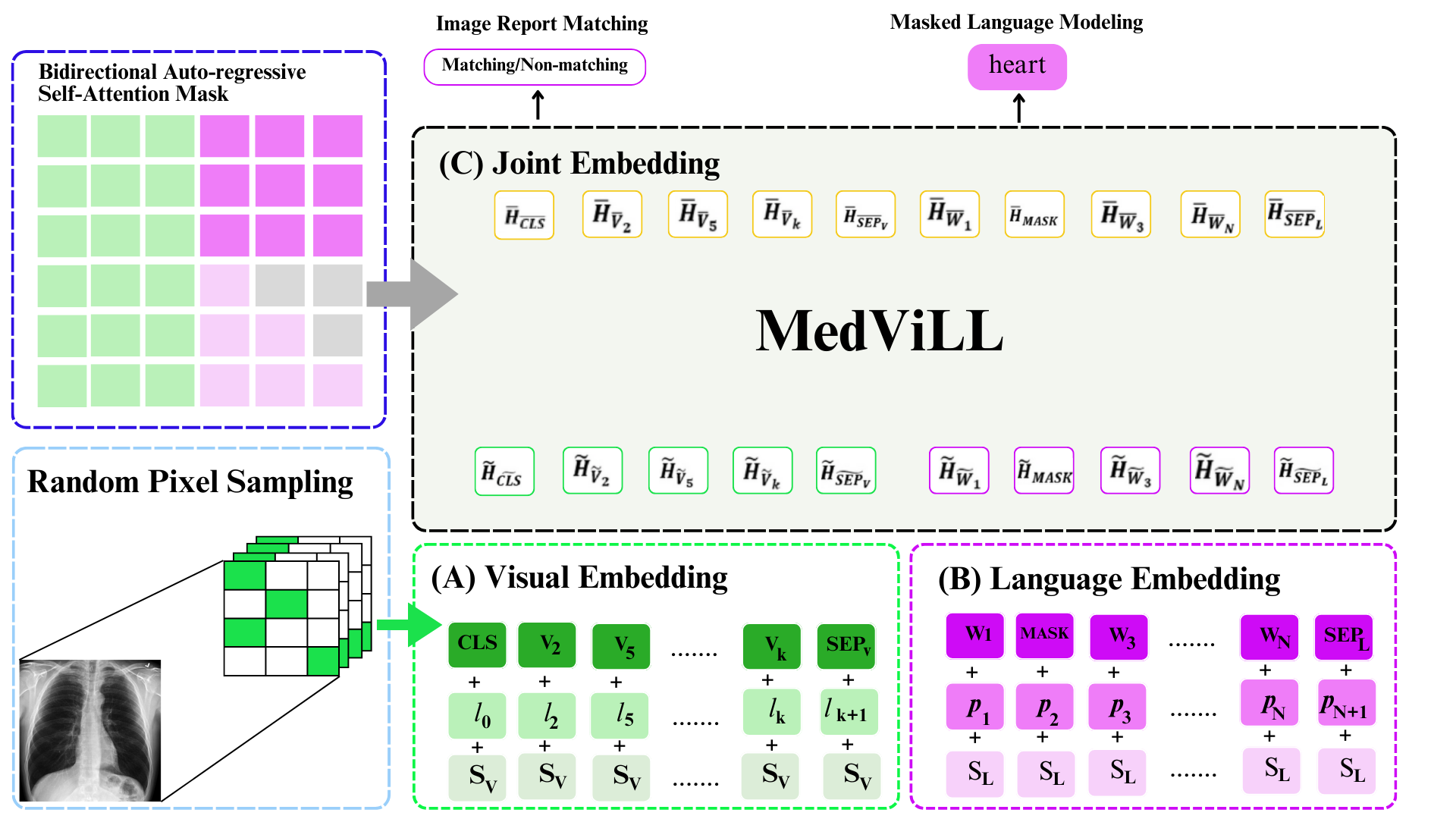}
    \caption{\textbf{Architecture of MedViLL \cite{Johnson2019}}. The model combines visual and language embeddings to enable joint representation learning for medical applications. (A) Visual embeddings are generated using random pixel sampling and positional encodings from medical images (e.g., X-rays). (B) Language embeddings incorporate tokens with segment and positional encodings from corresponding reports. Both embeddings are processed in (C) a joint embedding space through a bidirectional auto-regressive self-attention mechanism within a transformer. The model supports two primary tasks: image-report matching and masked language modeling, enabling robust multimodal understanding for clinical applications.}
    \label{fig:MedViLL}
\end{figure}

As illustrated in figure \ref{fig:MedViLL}, The visual embedding consist of extracting features from images using a CNN (ResNET-50). Practically, if $v$ is the flattened feature obtained from the CNN and $l$ the location feature, the final visual feature embedding is $v+l+s_v$ where $s_v$ is the semantic vector shared by all visual feature to differentiate themselves from language embedding. For language feature embedding, BERT \cite{devlin2019bertpretrainingdeepbidirectional} is used. Visual embedding and language embedding are concatenated to form the input of the joint embedding. 

The pre-training step of MedViLL consist of minimizing the negative log-likelihood:
\begin{equation}
    \mathcal{L}_{MLM}(\theta) = - \mathbf{E}_{(v,w) \sim D} \left[ log P_{\theta}(w_m|v,w_m) \right]
\end{equation}
Where $\theta$ is the trainable parameters, $(u,v)$ is a pair of images and its corresponding report. The pre-trained MedViLL achieved the score of complexity of 4.185, 84\% of accuracy and 6.6\% of BLUE4

\subsubsection{\textbf{MedCLIP}}
\begin{figure}[ht]
    \centering
    \includegraphics[width=\linewidth]{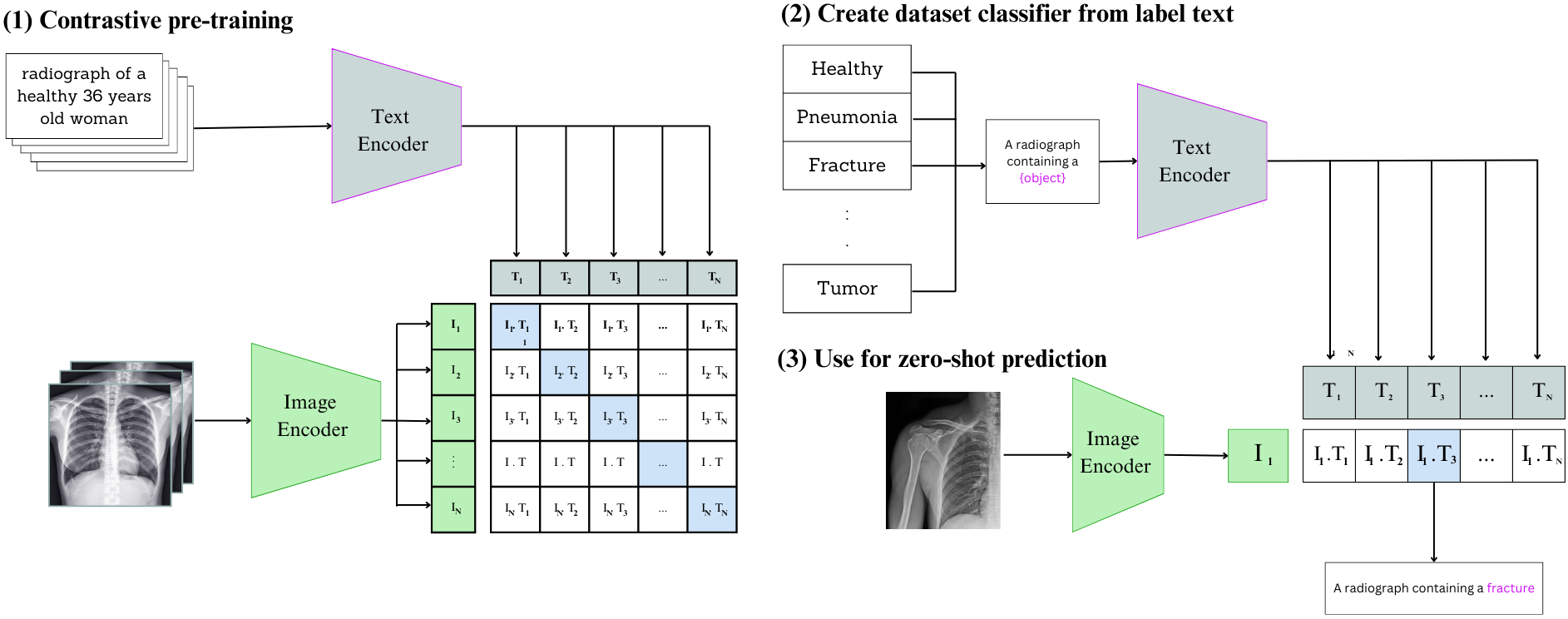}
    \caption{\textbf{Summary of the approach for CLIP \cite{radford2021learningtransferablevisualmodels}}: Contrastive pre-training aligns image and text embeddings (1), enabling the creation of dataset classifiers from textual labels (2), and facilitating zero-shot predictions by matching image embeddings with textual prompts (3). }
    \label{fig:CLIP}
\end{figure}

Several visual language models have gained recognition for their potential in healthcare. Contrastive Language-Image Pre-training (CLIP) \cite{radford2021learningtransferablevisualmodels}  is a powerful open-source model that has shown strong performance on image-text tasks in healthcare, including classifying medical images and generating radiology reports. As illustrated in figure \ref{fig:CLIP}, the contrastive pre-training consists of using ResNet or Vision Transformer(ViT) to extract features from images. The ResNet versions include improvements like anti-aliasing and attention pooling. The text encoder utilizes a Transformer model to convert text into feature representations. It processes text using a byte pair encoding (BPE) with a large vocabulary. The contrastive learning is then the process of learning to predict which text matches which image by maximizing the cosine similarity of correct (image, text) pairs and minimizing it for incorrect pairs. When testing,  the model can classify images into new categories by embedding the names or descriptions of the target classes and comparing them to the image embeddings.

MedCLIP \cite{wang2022medclipcontrastivelearningunpaired} has extended the capabilities of CLIP to the medical domain. Unlike traditional methods that rely on paired image-text datasets, MedCLIP decouples images and texts, significantly increasing the amount of usable training data. It addresses the issue of false negatives by incorporating medical knowledge to create a semantic matching loss, ensuring that images and reports with similar medical meanings are correctly identified. MedCLIP demonstrates superior performance in zero-shot prediction, supervised classification, and image-text retrieval tasks, outperforming state-of-the-art methods with much less pre-training data (20k data). This approach promises to improve the efficiency and accuracy of medical image analysis, supporting better clinical decision-making.

\subsubsection{\textbf{BioViL \& BioMedCLIP}}
BioViL (Biomedical Vision-Language Model)\cite{Boecking2022} is a specialized adaptation of CLIP, designed specifically for biomedical tasks involving medical images and text. Its architecture features a CNN image encoder, which generates a grid of local image embeddings, and a text encoder known as CXR-BERT, optimized for radiology report processing. Like CLIP, BioViL fuses image and text modalities into a shared latent space to create joint representations. It is trained on the MIMIC-CXR dataset with data augmentation. A key advantage of BioViL is its reduced reliance on extensive text-prompt engineering, enhancing usability in zero-shot classification tasks. BioViL demonstrates superior performance across multiple downstream tasks compared to other state-of-the-art models, achieving improved segmentation results without needing additional local loss terms or separate object detection networks. In addition, BioViL introduces the MS-CXR dataset, designed for evaluating image-text understanding in the radiology domain, which enhances the availability of resources for future studies.

Similar to MedCLIP, BioMedCLIP\cite{zhang2024biomedclipmultimodalbiomedicalfoundation} uses medical image-text pairs and applies contrastive learning to align the visual and textual modalities. It improves over general CLIP models by incorporating domain-specific medical knowledge.

\subsubsection{\textbf{ConVIRT}}
Contrastive Vision-Language Pre-training (ConVIRT) \cite{zhang2022contrastivelearningmedicalvisual} 
leverages a contrastive learning framework to pre-train models on large-scale medical image-text pairs. It is an unsupervised strategy to learn medical visual representations by exploiting
naturally occurring paired descriptive text. The framework takes input a pair of images $x_v$ and the text sequence $x_u$ to describe the imaging information. The goal is to learn an image encoder $f_v$ (modelled as a CNN) and then transfer it into downstream tasks. A random view $\tilde{x}_v$  is sampled from the inputs using a transformation function from a family of stochastic image transformations. A span $\tilde{x}_u$ is also sampled. Both $\tilde{x}_v$ and $\tilde{x}_u$ are encoded in a fixed-dimensional vector followed by a projection transforming $\tilde{x}_v$ and $\tilde{x}_u$ into $v$ and $u$ respectively. The training step involve two loss: the first is an image-to-text contrastive loss for a pair $i$
\begin{equation}
    l_i^{(v\rightarrow u)} = -log\frac{exp(\langle v_i, u_i \rangle/\tau)}{\sum_{k=1}^{N} exp(\langle v_i, u_k \rangle/\tau)}
\end{equation}

the second is the text-to-image contrastive loss

\begin{equation}
    l_i^{(u\rightarrow v)} = -log\frac{exp(\langle u_i, v_i \rangle/\tau)}{\sum_{k=1}^{N} exp(\langle u_i, v_k \rangle/\tau)}
\end{equation}
The final training loss is then a weighted combination of these two losses averaged over all positive image-text pairs in each minibatch.

ConVIRT is pre-trained on the second version of the public MIMIC-CXR dataset \cite{Johnson2019} and additional image-text pairs collected from the Rhode Island Hospital system. The pre-trained model is evaluated for three medical imaging tasks: image classification, zero-shot image-image retrieval, and zero-shot text-image retrieval. On four medical image classification tasks and two image retrieval tasks, ConVIRT outperformed other prominent in-domain initialization methods, resulting in representations of significantly higher quality. In comparison to ImageNet pretraining, ConVIRT achieved comparable classification accuracy while requiring an order of magnitude less labeled data.

\subsubsection{\textbf{VividMed}}
Previous VLMs have shown their capability to process medical visual data. However, these models typically rely on a single method of visual grounding and are limited to processing only 2D images. This approach falls short in addressing the complexity of many medical tasks, which require more versatile techniques, especially since a significant portion of medical images are 3D. Additionally, the scarcity of medical data further challenges these models. To overcome these limitations, VividMed (Vision Language Model with Versatile Visual Grounding for Medicine)\cite{luo2024vividmedvisionlanguagemodel} has been proposed as a more adaptable solution.

In VividMed, the task is not only to generate responses based on the input image and language instructions but also to identify key phrases $\{r_i\}_{i=1}^k$ in the generated text that refer to ROIs in the image.

\begin{figure}[ht]
    \centering
    \includegraphics[width=\linewidth]{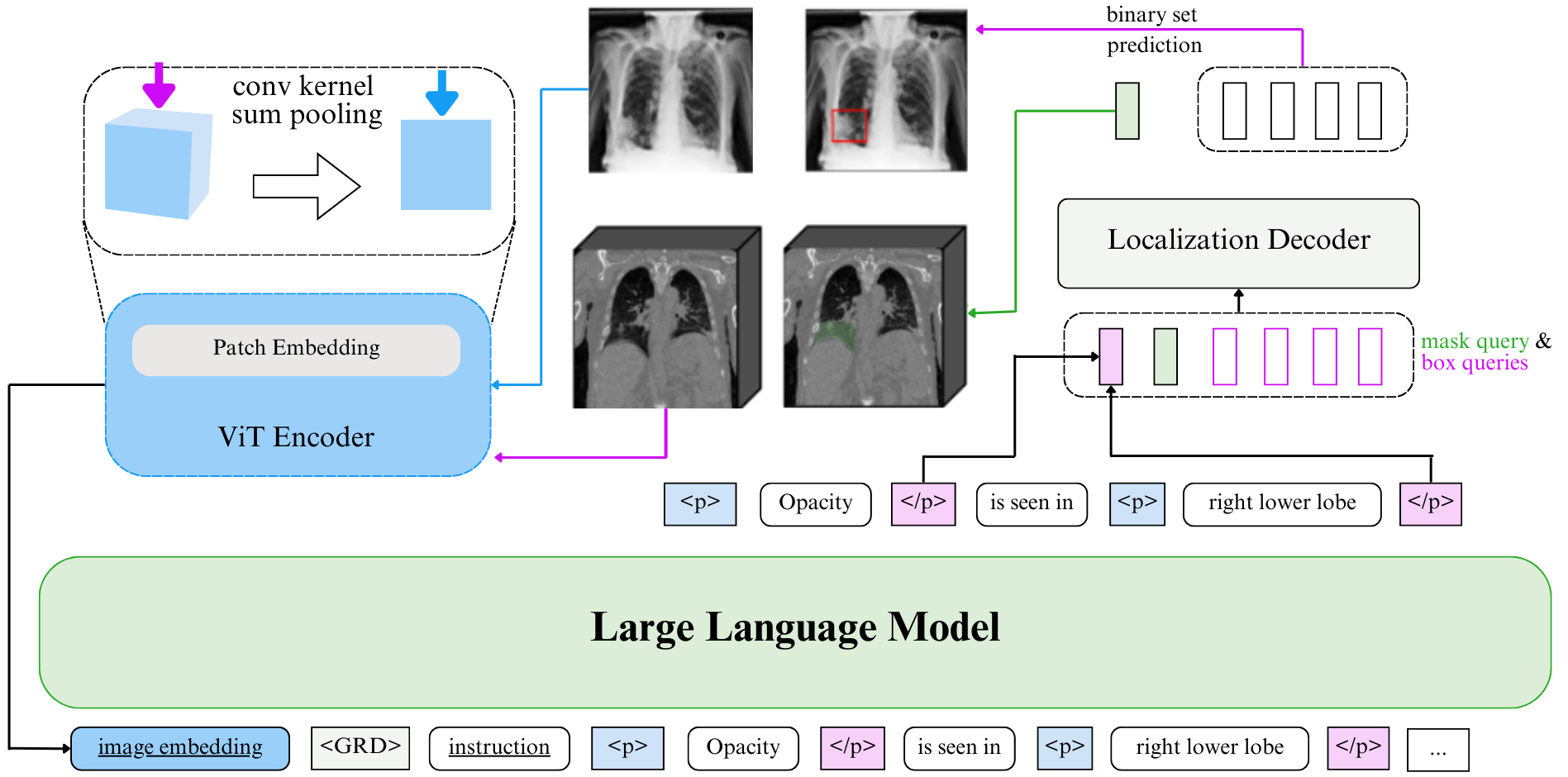}
    \caption{\textbf{Architecture of VividMed\cite{luo2024vividmedvisionlanguagemodel}}: Combines a ViT encoder for image embedding and a localization decoder for binary set prediction and spatial region identification, while leveraging a Large Language Model to generate medical descriptions based on multimodal inputs, including mask and box queries.}
    \label{fig:VividMed}
\end{figure}

The architecture of VividMed, shown in Figure \ref{fig:VividMed}, is built on CogVLM \cite{wang2024cogvlmvisualexpertpretrained} as the base VLM, designed to generate responses based on input images and language instructions.  Special tokens such as \verb|<p>|, \verb|</p>|, \verb|<grd>| and \verb|</grd>| are incorporated to specify the target phrase for grounding and indicate when the model should perform visual grounding. These tokens enable more precise control over the grounding process.

The promptable localization module follows Segment Anything Model (SAM) \cite{kirillov2023segment}, consisting of a vision encoder and a transformer-based decoder. To ground each phrase identified by the VLM, an embedding is generated by extracting the last-layer hidden state of the closing token (\verb|</p>|), which is then passed through an MLP to serve as a prompt for the decoder. The decoder generates bounding boxes or segmentation masks based on this prompt and the encoded image.

However, adapting the SAM mask decoder to output bounding boxes is complex. SAM's mask decoder produces a single binary mask per prompt, which cannot capture multiple instances of the same phrase. A workaround, like merging bounding boxes, results in information loss.

To address this, the authors introduce a new branch in the decoder for instance-specific predictions, inspired by DETR-like methods \cite{carion2020endtoendobjectdetectiontransformers}. They add multiple query tokens, each potentially linked to a unique instance or a dummy negative. The set of ground-truth labels and predictions is padded with dummy negatives to match a predefined token count ( $m$), ensuring coverage of different instances.

During training, the Hungarian algorithm assigns each prediction a unique label to minimize a cost function, $( L_{\text{cost}} )$, which combines a bounding box regression loss $( L_{\text{box}} )$ and a discrimination loss $( L_{\text{disc}} )$. The authors employ $( \ell_1 )$ and GIoU losses for $( L_{\text{box}} $) and focal loss for $( L_{\text{disc}} )$, as used in DINO \cite{zhang2023dino}. For segmentation masks, the loss combines Dice and focal losses.

Since most medical images consist of multiple 2D slices stacked to form a 3D structure, a single-slice X-ray represents a simpler case. To avoid unwanted artifacts from inter-slice interpolation in 3D images, the authors dynamically adjusted model weights based on the image slice count, following methods for universal backbones in medical imaging \cite{luo2024buildinguniversalfoundationmodels}. In the ViT-based vision encoder (Fig. \ref{fig:VividMed}), a maximum patch number $( t_d )$ and base patch size $( P_d )$ are set along the depth, adjusting the effective patch size $( P'_d )$ dynamically according to the input slices. For upsampling, transposed convolutions with a scale factor of 2 are used, disabling upsampling along the depth when feature maps reach the input depth, with kernel weights adjusted by mean pooling to ensure size consistency.

The training procedure of VividMed is conducted in 3 stages: (i) visual grounding pre-training, the model is instructed here to determine whether given targets detection exist on the image and list the target names along with their presence in the respons; (ii) medical visual instruction tuning, dedicated to training the model’s visual understanding and reasoning capabilities for medical images; (iii) alignment, consisting of finetuning the model to align both the visual grounding and medical image understanding abilities
trained by previous stages to unleash the combined strengths.

VividMed is trained on VinDr-CXR, MIMIC-CXR, and CT-RATE for the task of VQA; ROCOv2\cite{R_ckert_2024} dataset for Image Captioning; MIMIC-CXR\cite{Johnson2019} and CT-RATE\cite{hamamci2024developinggeneralistfoundationmodels} for Report Generation. The model is evaluated using BLEU, ROUGE and METEOR over VQA-RAD\cite{Lau2018}, SLAKE\cite{liu2021slakesemanticallylabeledknowledgeenhanceddataset}, VQA-Med\cite{Ben2019} (for VQA tasks) and the test sets of MIMIC-CXR and CT-RATE for Report Generation. As results, ividMed shows non-trivial general improvement over fine-tuned CogVLM and outperforms all other baselines in VQA. For Report Generation, VividMed outperforms all other baselines by a large margin on both datasets.

\subsubsection{\textbf{Flamingo-CXR \& Med-Flamingo}}
\begin{figure}[ht]
    \centering
    \includegraphics[width=1\linewidth]{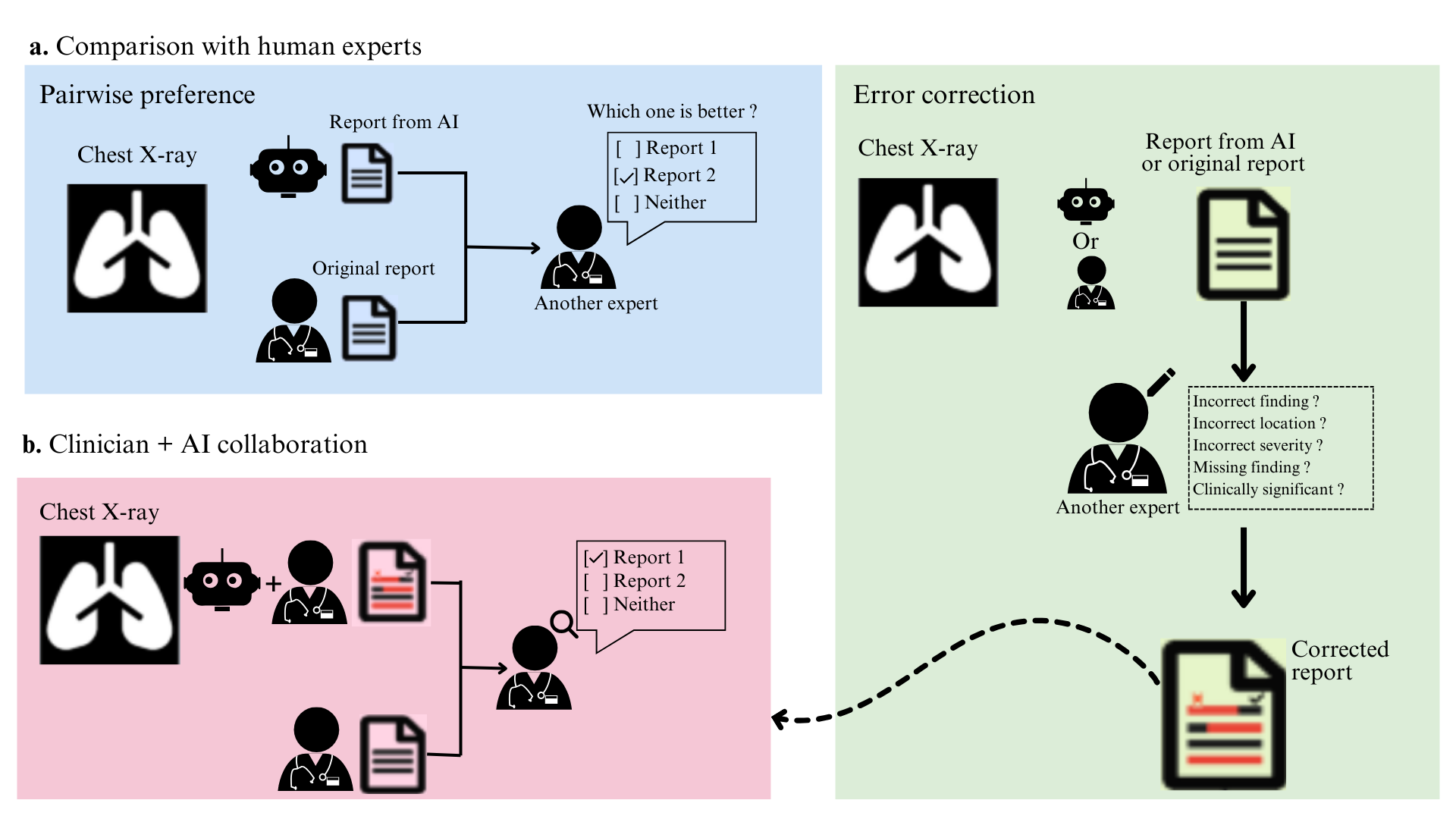}
    \caption{\textbf{Overview of the human evaluation framework by \cite{Tanno2024}}. \textbf{(a)} Two  evaluation methods for comparing radiology reports from an AI model with those created by human experts. The first method involves a pairwise preference test where an expert chooses which report—AI-generated or human-written—is better for patient care. The second method is an error correction task, where the expert reviews a single report, makes edits, and explains their significance. Additionally, \textbf{(b)} there is an assessment of the AI's effectiveness in an assistive role by comparing an edited AI report to a human-only report, using data from outpatient care in India and intensive care in the U.S., with board-certified radiologists evaluating regional differences.}
    \label{fig:Flamingo-CXR}
\end{figure}

Flamingo-CXR\cite{Tanno2024} is a model designed to generate radiology reports for Chest X-rays. As many VLMs, Flamingo-CXR aims to improve patient care and reduce radiologists' workload by automating report generation. It addresses the challenge of evaluating the clinical quality of AI-generated reports by engaging a panel of board-certified radiologists for expert evaluation. 
It was fine-tuned using two large datasets (MIMIC-CXR from a US emergency department and IND1 from in/outpatient settings in India). The evaluation involved comparing AI-generated reports with human-written ones through a pairwise preference test and an error correction task. Radiologists assessed the reports' clinical quality as illustrated in Fig. \ref{fig:Flamingo-CXR}, identifying errors and providing corrections. The model showed that 56.1\% of Flamingo-CXR reports for intensive care were preferable or equivalent to clinician reports, rising to 77.7\% for in/outpatient X-rays and 94\% for cases with no pertinent abnormalities. Both human and AI-generated reports contained errors, with 24.8\% of in/outpatient cases having clinically significant errors in both types. The authors found that clinician-AI collaboration improved report quality, with AI-generated reports corrected by experts being preferable or equivalent to original clinician reports in 71.2\% of IND1 cases and 53.6\% of MIMIC-CXR cases.

Med-Flamingo\cite{moor2023medflamingomultimodalmedicalfewshot}, for its part, is based on OpenFlamingo-9B and is pre-trained on paired and interleaved medical image-text data from publications and textbooks. The training involved constructing a unique dataset from over 4,000 medical textbooks and the PMC-OA dataset, resulting in a comprehensive collection of medical images and text. The model was trained using multi-GPU setups and advanced optimization techniques to handle the complexity and multimodality of medical data. It achieved a great performance in medical VQAs, up to 20\% improvement in clinician ratings over previous models. The model was evaluated on several datasets, including the Visual USMLE dataset, and showed superior performance in generating clinically useful answers. The human evaluation study with clinical experts confirmed that Med-Flamingo’s answers were most preferred by clinicians, highlighting its potential for real-world medical applications.

\subsubsection{\textbf{BiomedGPT}}
BiomedGPT\cite{Zhang2024} is an open-source model, lightweight vision–language foundation model designed to perform diverse biomedical tasks. Unlike traditional AI models that are specialized for specific tasks, BiomedGPT aims to be a generalist, capable of interpreting different data types and generating tailored outputs. This model addresses the limitations of existing biomedical AI solutions, which are often heavyweight and closed-source, by offering a versatile and accessible alternative.

BiomedGPT was pretrained using a large and diverse dataset comprising 592,567 images, approximately 183 million text sentences, 46,408 object-label pairs, and 271,804 image-text pairs. The model was fine-tuned on various multimodal tasks, including visual question answering (VQA) and image captioning, using datasets like VQA-RAD, SLAKE, and PathVQA. The performance of BiomedGPT was benchmarked against leading models using recognized metrics from the literature, and human evaluations were conducted to assess its capabilities in radiology visual question answering, report generation, and summarization.

BiomedGPT achieved state-of-the-art results in 16 out of 25 experiments while maintaining a computing-friendly model scale. It demonstrated robust prediction ability with a low error rate of 3.8\% in question answering, satisfactory performance with an error rate of 8.3\% in writing complex radiology reports, and competitive summarization ability with a nearly equivalent preference score to human experts. The study highlights BiomedGPT’s potential in improving diagnosis and workflow efficiency in medical applications, although further enhancements are needed for clinical usability.

\subsubsection{\textbf{RadFM}}
The Radiology Foundation Model (RadFM)\cite{wu2023generalistfoundationmodelradiology} is designed to handle a wide range of clinical radiology tasks by learning from 2D and 3D medical scans and corresponding text descriptions. The authors constructed a large-scale Medical Multi-modal Dataset (MedMD), consisting of 16 million radiology scans and high-quality textual descriptions. The model was pre-trained on MedMD and fine-tuned on a subset called RadMD, which includes 3 million meticulously curated multi-modal samples. The evaluation of RadFM was conducted using RadBench, a comprehensive benchmark covering tasks like disease diagnosis, report generation, and visual question answering. It demonstrated significant improvements across all evaluated tasks compared to existing models like OpenFlamingo, MedVInT, MedFlamingo, and GPT-4V. It excelled in modality recognition, disease diagnosis, medical visual question answering (VQA), report generation, and rationale diagnosis. The model’s performance was validated through both automatic and human evaluations, showing superior results in terms of accuracy, precision, and recall. The study highlights RadFM’s potential to unify 2D and 3D radiologic images and support multiple medical tasks, marking a significant step towards developing a generalist foundation model for radiology.

\subsubsection{\textbf{LLaVa-Med}}
LLaVa-Med\cite{li2023llavamedtraininglargelanguageandvision} enhances medical VQA through the integration of vision-language models. The authors developed a structured approach that involves curating high-quality instruction-following data based on medical images and their corresponding textual descriptions. This process includes filtering a large dataset to focus on single-plot images across common imaging modalities, such as chest X-rays, CT scans, and MRIs. They manually curated few-shot examples to demonstrate effective conversation generation based on provided captions and contextual information extracted from PubMed papers. This structured approach aims to improve the model's ability to generate accurate and contextually relevant responses to visual questions in the medical domain. 
Quantitative results showcased significant improvements in accuracy for both open-ended and closed-ended questions compared to prior methods. For instance, their model variants demonstrated high recall rates and accuracy across various question types, indicating enhanced generalization capabilities. Notably, the LLaVA-Med model achieved new state-of-the-art results, underscoring the effectiveness of their instruction-following data generation strategy and the model's ability to integrate external knowledge for better contextual understanding.

\subsubsection{\textbf{MedVInT}}
The MedVInT\cite{zhang2024pmcvqavisualinstructiontuning} model utilizes a generative learning approach for Medical Visual Question Answering (MedVQA), achieved by aligning a pre-trained vision encoder with a large language model through visual instruction tuning. The model is pre-trained on the proposed PMC-VQA\cite{zhang2024pmcvqavisualinstructiontuning} dataset, which covers various medical modalities and diseases, significantly exceeding the size and diversity of existing datasets. The training process involves fine-tuning the model on established benchmarks such as VQA-RAD\cite{Lau2018} and SLAKE\cite{liu2021slakesemanticallylabeledknowledgeenhanceddataset}, leading to substantial performance improvements. Notably, two versions of MedVInT, named MedVInT-TE and MedVInT-TD, are tailored for different types of questions—open-ended and close-ended, respectively. The model's architecture is designed to leverage domain-specific pre-training, enhancing its ability to interpret medical visuals accurately and generate relevant answers to text-based queries.
For open-ended questions, the accuracy improved significantly, with MedVInT-TE achieving a 16\% increase on VQA-RAD and 4\% on SLAKE when pre-trained. In terms of specific results, the MedVInT-TE version reached an accuracy of 70.5\% on the ImageCLEF\cite{ImageCLEF2024} benchmark, surpassing the previous state-of-the-art accuracy of 62.4\%. Additionally, when compared to models trained from scratch, those utilizing the PMC-CLIP vision backbone showed consistent performance improvements, underscoring the importance of domain-specific pre-training. Overall, the MedVInT model outperformed several existing models, including LLaVA-Med, demonstrating its effectiveness in handling both open-ended and close-ended questions in medical image interpretation.

\subsection{VLMs integration in the medical workflow}
The practical application of VLMs in the medical domain is diverse and profound. VLMs offer capabilities such as multimodal data interpretation, where they integrate textual data (e.g., patient records) with visual inputs (e.g., medical images), enabling enhanced diagnostic and decision-making processes. For instance, these models can assist in generating detailed reports from radiological images or identifying anomalies in pathology slides. Such tools have the potential to improve the efficiency and accuracy of diagnostic workflows, especially in resource-constrained settings \cite{bedi_liu_2024, yiving_jesutofummi_2024}.

In clinical decision support systems, VLMs can interpret complex datasets to predict patient outcomes or recommend treatments. Models like GPT-4V have been used to evaluate dermatological conditions, providing differential diagnoses and triaging recommendations. However, their integration into clinical workflows requires addressing challenges such as bias in predictions, lower performance in rare conditions, and the need for robust privacy protections.

Another critical application is in medical education and training. VLMs can synthesize multimodal datasets to create case simulations for trainees, facilitating experiential learning. For instance, generating synthetic but realistic case scenarios from de-identified datasets can provide students and clinicians with valuable practice.

Despite their promise, integrating VLMs into the medical workflow requires rigorous validation and regulatory approval. Ethical considerations, such as ensuring equity in model performance across diverse patient populations, must also be addressed to avoid perpetuating existing disparities in healthcare access and quality.

% You will need to summarize the datasets used in these fields: for instance Medtrinity-25M, GAIM-MMbenchmark, etc.

% Summarize RSNA Pneumonia Detection\cite{Wang_2017}, CheXpert\cite{irvin2019}, COVIDx\cite{wang2020covidnettailoreddeepconvolutional} and MURA\cite{rajpurkar2018muralargedatasetabnormality}. 

% Also talking about how they used to evaluate the model. Benchmarks, Meterics, etc

\section{VLM benchmarking and Evaluations}
% What are the most well-known models in this field: both open-source and close-source models
The progress of VLMs in medicine heavily depends on the availability of high-quality datasets. These datasets, along with established benchmarks, have played a pivotal role in driving both research and practical applications in the field. Several key datasets and benchmarks have already contributed significantly to the development and evaluation of VLMs, shaping the advancements in this domain.

\begin{table*}[ht]
    % \centering
    \caption{\textbf{Benchmarking in Medical VLMs}. This table summarizes key benchmarks in medical VLMs, including their modalities, dataset sizes, tasks, and sources. The benchmarks cover a wide range of medical imaging modalities and tasks, such as classification, segmentation, and VQA}
    \label{tab:benchmarking }
    \begin{tabular}{lllll}
        \toprule
        \textbf{Benchmark} & \textbf{Modality} & \textbf{Size} & \textbf{Task} & \textbf{Source} \\
        \midrule
        CheXpert\cite{irvin2019} & 1 & 224,316 & 14 (Pathology classification tasks) & from 65,240 patients \\
        CheXpert Plus\cite{chambon2024chexpertplusaugmentinglarge} & 1 & 224,316 & 14 (Classification + Localization tasks) & from 187,711 studies and  64,725 patient \\
        MIMIC-CXR-JPG\cite{johnson2019mimiccxrjpglargepubliclyavailable} & 2 & 377,110 & 14 (Pathology classification + Report generation) & from 227,827 imaging studies \\
        Medtrinity-25M\cite{xie2024medtrinity25mlargescalemultimodaldataset} & 3 & 25M & 30+ (Multi-task learning)& from radiology, pathology, and EHRs$^\dagger$ \\
        GMAI-MMBench\cite{chen2024gmaimmbenchcomprehensivemultimodalevaluation} & 38 & 26k & 18 (Diverse medical tasks) & 28 datasets from hospitals \\
        PMC-VQA\cite{zhang2024pmcvqavisualinstructiontuning} & 2 & 1.5M & 4 (Visual Question Answering tasks) & PubMed Central \\
        PathVQA\cite{he2020pathvqa30000questionsmedical} & 1 & 6k & 7 (Visual Question Answering tasks) & Textbook, PEIR \\
        VQA-RAD\cite{Lau2018} & 3 & 3k & 11 (Visual Question Answering tasks) & Teaching cases from Medpix \\
        BESTMVQA\cite{hong2023bestmvqabenchmarkevaluationmedical} & 2 & N/A & 5 (Visual Question Answering tasks) & as X-rays, CT scans, MRIs, and pathology slides. \\
        CT-RATE\cite{hamamci2024developinggeneralistfoundationmodels} & 2 & N/A & 3 (Classification, Segmentation tasks) & from radiology departments \& imaging repositories. \\
        SLAKE\cite{liu2021slakesemanticallylabeledknowledgeenhanceddataset} & 3 & 2k & 10 (Visual Question Answering tasks) & MSD, Chesx-ray8, CHAOS \\
        RadBench\cite{kuo2024radbenchevaluatinglargelanguage} & 6 & 137k & 5 (Image-text understanding tasks) & 13 image-text paired datasets \\
        OmniMedVQA\cite{hu2024omnimedvqanewlargescalecomprehensive} & 12 & 128k & 5 & 73 classification datasets\\
        \bottomrule
    \end{tabular}
    $\dagger$: Electronic Health Records (EHRs)
\end{table*}

\subsection{CheXpert \& CheXpert Plus}
Chest X-rays (CXR)\cite{Wang_2017} are one of the most commonly used imaging techniques in medical diagnostics. They provide a non-invasive and quick method to capture images of the lungs, heart, airways, blood vessels, and bones of the chest. CXRs are essential for diagnosing a variety of conditions, including:
pulmonary diseases \footnote{Pulmonary diseases include pneumonia, tuberculosis, and lung cancer},
cardiac conditions \footnote{Cardiac conditions include heart failure and cardiomegaly},
trauma \footnote{Trauma include fractures of the ribs or injuries to the lungs} and
infections \footnote{Infections related to bronchitis or pleural effusion (fluid in the lungs).}

Due to their widespread use in medical practice, chest X-rays generate a vast amount of clinical data, making them an ideal modality for training machine learning models. This volume of data, coupled with the relative simplicity of interpreting certain conditions on X-rays, makes chest X-rays a prime focus for AI-based research and development, including in the area of VLMs.

CheXpert \cite{irvin2019} is a large-scale dataset widely used in medical imaging research, particularly for chest X-ray analysis. Developed by researchers at Stanford University, it serves as a valuable resource for training and evaluating machine learning models in radiology. The dataset is notable for its size, diversity, and comprehensive labeling of medical conditions observed in chest X-rays. However, CheXpert has certain limitations, most notably the lack of demographic information, which can be critical for developing equitable and generalizable AI models.

To address these limitations, CheXpert Plus \cite{chambon2024chexpertplusaugmentinglarge} was introduced. This enhanced dataset includes 36 million tokens comprising images, radiology reports, demographic details, 14 pathology labels, RadGraph annotations, and pre-trained models for key machine learning tasks. By incorporating these additional elements, CheXpert Plus significantly expands the scope and utility of the original dataset, enabling more robust and inclusive research in medical AI.

\subsection{MIMIC-CXR-JPG}
The MIMIC-CXR-JPG \cite{johnson2019mimiccxrjpglargepubliclyavailable} dataset is a large collection of chest radiographs designed to facilitate research in automated analysis of chest images. It consists of 377,110 chest X-ray images associated with 227,827 imaging studies collected from the Beth Israel Deaconess Medical Center between 2011 and 2016. The dataset includes 14 binary labels indicating the presence or absence of various pathologies derived from NLP tools applied to radiology reports.

\subsection{Medtrinity-25M}
\begin{figure}[ht]
    \centering
    \includegraphics[width=\linewidth]{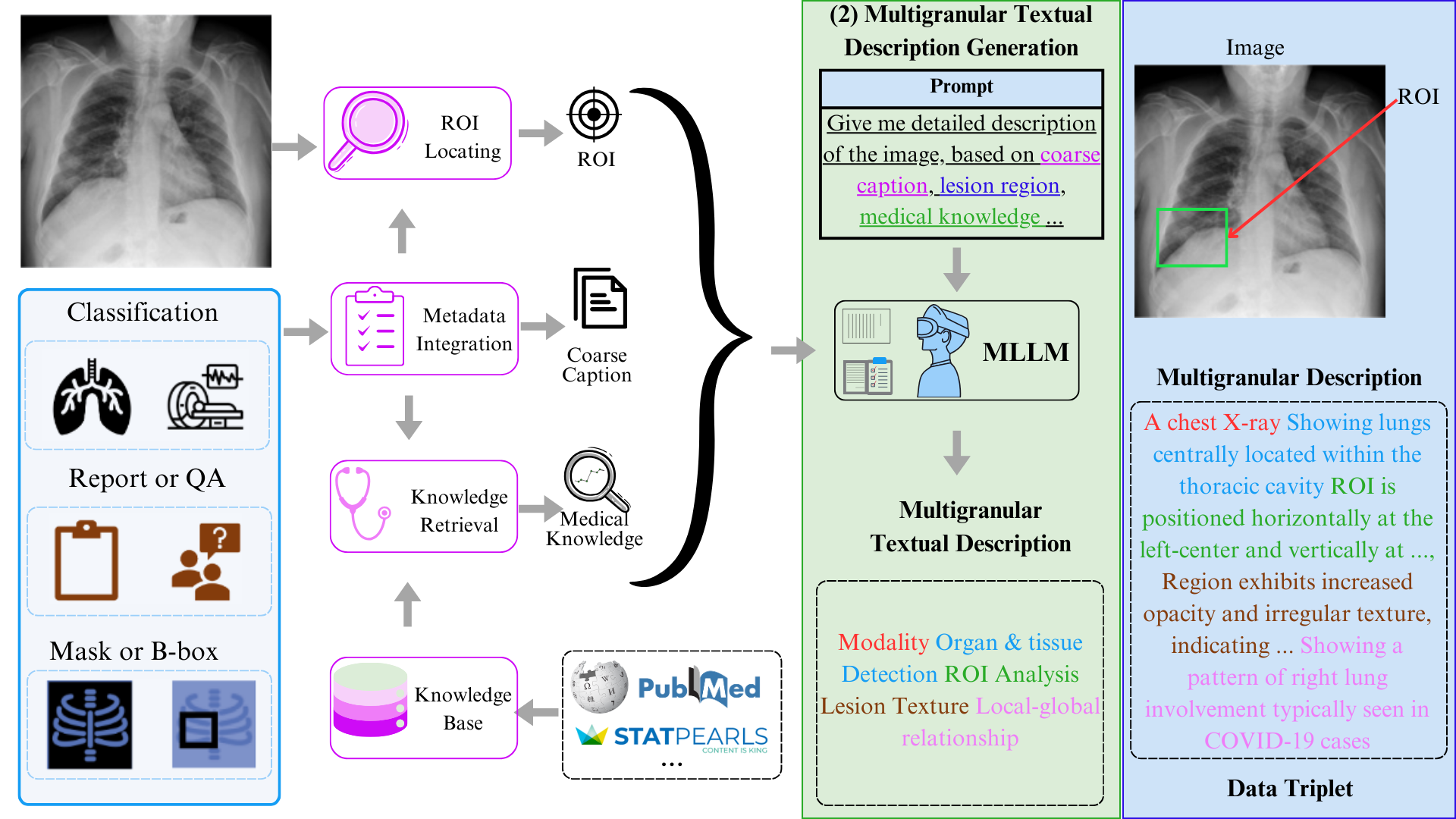}
    \caption{\textbf{Data pipeline of Medtrinity-25M\cite{xie2024medtrinity25mlargescalemultimodaldataset}}. The framework consists of three main components:  (1) Data processing pipeline that integrates ROI localization, metadata integration, and medical knowledge retrieval from established databases like PubMed and STATPEARLS, (2) Multigranular textual description generation powered by MLLM incorporating prompts based on coarse captions and medical knowledge, and (3) Data triplet output displaying the original image with ROI annotation alongside comprehensive multigranular descriptions. The framework leverages classification, report analysis, and knowledge integration to generate detailed medical image interpretations.}
    \label{fig:medtrinity-25m}
\end{figure}

Medtrinity-25M\cite{xie2024medtrinity25mlargescalemultimodaldataset} is a large-scale multimodal dataset with multigranular annotations for medicine. The dataset is a set of \verb|{image,ROI,description}| where \verb|ROI| (Region Of Interest) is associated with an abnormality and is represented by a bounding box or a segmentation mask, specifying the relevant region within the corresponding \verb|image|. Each of these images is associated with a multigranular textual \verb|description| describing the disease/lesion type and other discoveries as illustrated in the pipeline in Figure \ref{fig:medtrinity-25m}.

The images are collected from various sources including online resources (Kaggle, Zenodo, Hugging Face, GitHub,etc.), relevant medical dataset research (such as CheXpert). They contain either local or global annotations according to their sources, and 25,001,668 samples spanning 10 modalities and over 65 diseases have been collected. The captions and annotations like masks and bounding boxes from these sources are utilized to construct ROIs and corresponding textual descriptions.

As illustrated by Figure \ref{fig:medtrinity-25m}, the pipeline consists of two stages. The first stage is the data processing, where the lack of domain-specific knowledge is addressed by integrating metadata, ROI locating, and medical knowledge retrieval. These informations processed in the first stage are used in the second stage to prompt LLMs (GPT-4V, LLAVA-Med, LLaMA3) to produce the multigranular textual descriptions.

MedTrinity-25M being a large-scale dataset (25 million image-text pairs), is designed to advance research in medical AI,  providing a rich resource for various tasks in medical image analysis and natural language processing. It can be used to develop and fine-tune large multimodal VLMs capable of understanding both visual and textual medical data.

\subsection{GMAI-MMBench}
GMAI-MMBench\cite{chen2024gmaimmbenchcomprehensivemultimodalevaluation} is a comprehensive multimodal evaluation benchmark towards general medical AI. It features data from 284 datasets across 38 medical imaging modalities and spans 18 clinical-related tasks, 18 medical departments, and 4 perceptual granularities in VQA format. These datasets are organized in a way that allows for flexible and customizable evaluations through a lexical tree, making it easier for researchers to focus on specific medical AI challenges.

\begin{figure}[ht]
    \centering
    \includegraphics[width=\linewidth]{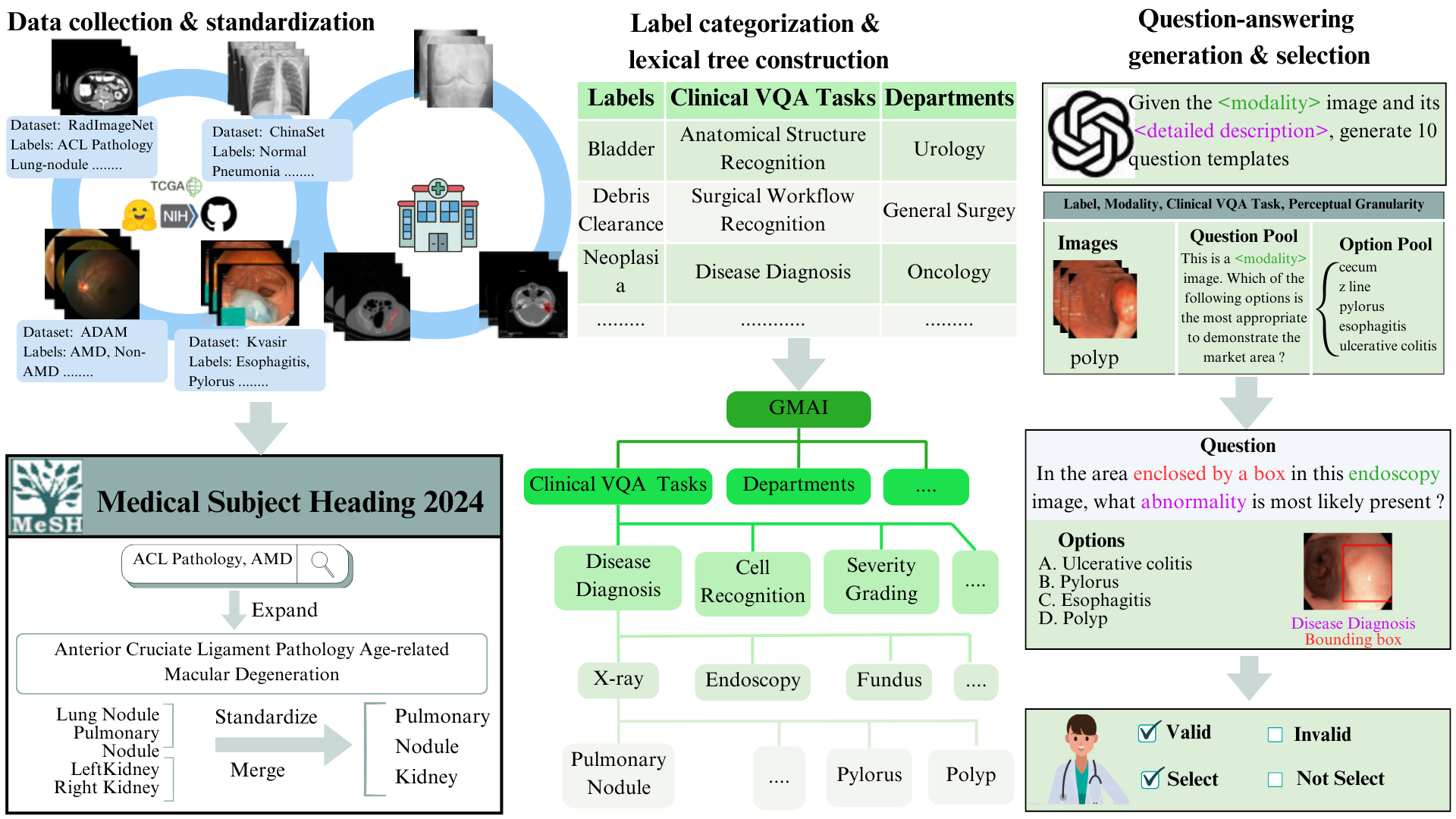}
    \caption{\textbf{Overview of the GMAI-MMBench\cite{chen2024gmaimmbenchcomprehensivemultimodalevaluation} framework} comprising three main components: (1) Data collection and standardization pipeline integrating multiple medical imaging datasets (RadImageNet, ChinaSet, ADAM, Kvasir) with standardized labeling through Medical Subject Heading 2024, (2) Label categorization and hierarchical lexical tree construction organizing clinical VQA tasks across medical departments, and (3) Question-answering generation and selection system producing structured questions with modality-specific options and validation criteria. The framework enables comprehensive evaluation of medical visual question-answering capabilities across different imaging modalities, clinical tasks, and diagnostic scenarios.}
    \label{fig:gmai-mmbench}
\end{figure}

The GMAI-MMBench benchmark is constructed in three steps as illustrated in Figure \ref{fig:gmai-mmbench}. The first step consists of data collection and standardization where the datasets are sourced both from Internet searches and hospitals. As standardization, all 2D/3D medical images are converted into 2D RGB images and some additional processing such as expanding all abbreviations,  unifying different expressions for the same target, and merging labels with left and right distinctions. The second step is the label categorization and lexical tree construction where three subjects tailored for the biomedical field are proposed from data; they include clinical VQA tasks, departments, and perceptual granularities. The well-organized structure from this categorization is then used in the third step to generate VQA pairs for labels. 

First, candidate questions are formulated by combining modality, clinical task hints, and perceptual granularity information. For each label, 10 options are selected at random, and GPT-4o is employed to generate 10 candidate questions for each, ensuring diversity in question phrasing and content. After this automated generation process, the questions undergo manual review by medical professionals to ensure relevance, accuracy, and clarity. For the options generation, the process is carefully designed to handle different levels of perceptual granularity in medical images, separating global (image-level) and local (mask, bounding box, or contour-level) views. For the global view, answer options are sourced from other categories in the dataset, ensuring no overlap with the correct answer to avoid ambiguity. In the local view, options are derived from a shared pool based on modality, clinical task, and the specific detail being queried. Second, a question is selected for each image, and a set of answer options (including the correct one) is randomly generated, ensuring a well-structured and diverse set of question-answer pairs for evaluation. Third, manual validation and selection are performed to ensure data quality and balanced distribution.

This benchmark has been used to evaluate several LVLMs, including general models and medical-specific models. 29 out of 50 models are picked and evaluated using macro-averaged accuracy (ACC) as the evaluation metric for single-choice questions.For multiple-choice questions, the models are evaluated using both macro-averaged accuracy ($\text{ACC}_{\text{mcq}}$) and recall ($\text{Recall}_{\text{mcq}}$) to account for the proportion of correct answers selected relative to the total predictions and the ground-truth options. This evaluation revealed that medical tasks are still challenging for all LVLMs, since even the most advanced model (GPT-4o) is limited to 54\% accuracy.

\subsection{PMC-VQA}
PMC-VQA (PubMed Central Visual Question Answering)\cite{zhang2024pmcvqavisualinstructiontuning} is a groundbreaking large-scale medical Visual Question Answering dataset, introduced by Zhang et al. in 2023. The dataset comprises 227,117 diverse image-question-answer triplets, extracted from over 40,000 medical research papers in PubMed Central. It features a wide range of medical images, including radiological scans, pathology slides, clinical photographs, and medical illustrations, paired with expert-authored questions and answers. The dataset's strength lies in its comprehensive coverage of various medical domains and question types, from diagnostic interpretations to anatomical identifications and technical analyses.

What sets PMC-VQA apart is its instruction tuning approach for medical visual understanding tasks. The dataset not only provides factual question-answer pairs but also incorporates complex reasoning scenarios that mirror real clinical decision-making processes. With a reported model performance achieving approximately 76\% overall accuracy and clinical correctness of 72\%, PMC-VQA has established itself as a benchmark for evaluating and advancing medical VQA systems. Its applications span clinical decision support, medical education, and research assistance, making it a valuable resource for developing more robust and clinically relevant medical AI systems.

\subsection{Hallucination benchmark}
The pretrained medical VLMs have shown great potentials for VQA tasks but are not tested on hallucination phenomenon in the clinical settings. To solve this problem, a hallucination benchmark in medical VQA \cite{wu2024hallucinationbenchmarkmedicalvisual} have been developed, aimed at evaluating the performance of LLMs in detecting hallucinations during medical VQA. It outlines the importance of minimizing hallucinations in healthcare settings, where inaccuracies could lead to misdiagnoses or inappropriate treatments. The authors modified existing VQA datasets (PMC-VQA, PathVQA and VQA-RAD) to assess model responses to both genuine and nonsensical questions, as well as mismatched image queries. They analyzed the performance of various models, including LLaVA and GPT-4, highlighting that while GPT shows strong overall performance, it poses privacy concerns in clinical settings. The findings suggest that the LLaVA-v1.5-13B model, when combined with an effective prompting strategy (L+D), exhibits the best accuracy and minimal irrelevant predictions, making it suitable for deployment as a visual assistant in healthcare environments.

\subsection{BESTMVQA}
The Benchmark Evaluation System for Medical Visual Question Answering (BESTMVQA) \cite{hong2023bestmvqabenchmarkevaluationmedical} involves several key components designed to address challenges such as data insufficiency and the need for a unified evaluation system. The process begins with data preparation, where users upload their self-collected clinical data. This data is then processed using a semi-automatic tool that extracts medical images and relevant texts for medical concept discovery. A human-in-the-loop framework assists in annotating these medical concepts, which are auto-labeled initially and later verified by professionals. The annotated data is utilized to generate high-quality question-answer pairs with the help of a pre-trained language model. Furthermore, BESTMVQA provides a comprehensive model library containing a variety of state-of-the-art models for users to select from, facilitating ease in evaluating different models against benchmark datasets. Users can conduct experiments with simple configurations, allowing the system to automatically train and evaluate models, ultimately generating comprehensive reports on their performance and applicability in medical practice.

\subsection{CT-RATE dataset}
\begin{figure}[ht]
    \centering
    \includegraphics[width=1\linewidth]{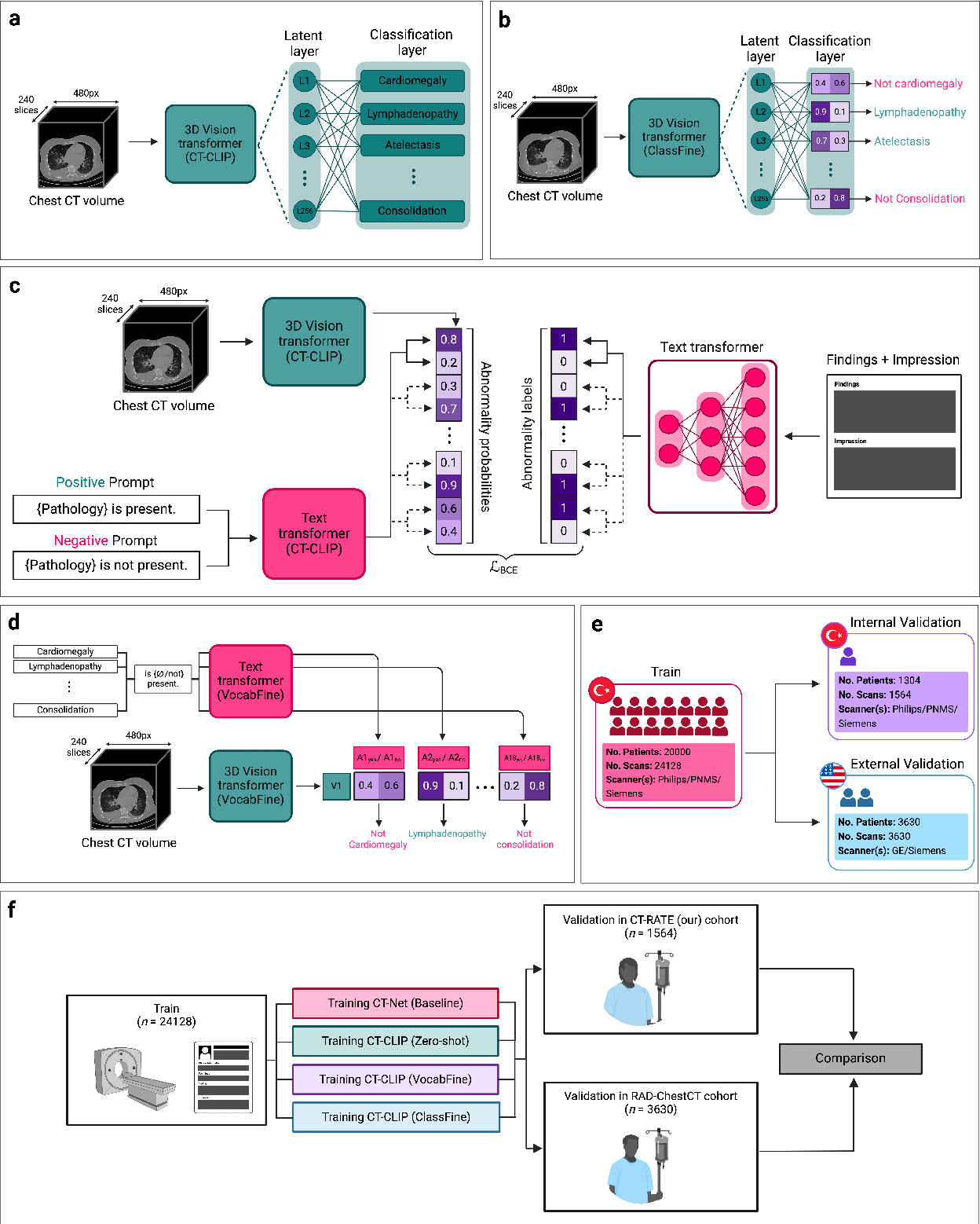}
    \caption{\textbf{Overview of the CT-RATE dataset and CT-CLIP \cite{hong2023bestmvqabenchmarkevaluationmedical}}: ( Finetuning CT-CLIP and the validation strategy. a. Illustration of linear probing finetuning method (ClassFine) for CT-CLIP, where a linear layer is incorporated into the vision encoder. b. ClassFine enables multi-abnormality classification but is limited to the classes predefined during finetuning. c. Illustration of CT-CLIP’s open vocabulary finetuning method (VocabFine) for each abnormality. d. VocabFine allows for open vocabulary abnormality classification even after finetuning, although it is constrained to the prompts provided during finetuning. e. Validation: models are trained on the CTRATE dataset and then tested on both an internal CT-RATE validation set and an external dataset ;  f. Comparison: a comprehensive evaluation is performed in multi-abnormality detection across two different cohorts, evaluating CT-CLIP, the two finetuned models, and a fully supervised method.}
    \label{fig:CT-RATE}
\end{figure}

The CT-RATE dataset \cite{hamamci2024developinggeneralistfoundationmodels} comprises pairs of 3D medical images and corresponding textual reports. Initially, it included 25,692 non-contrast 3D chest scans, which were subsequently expanded to over 14.3 million 2D slices using advanced reconstruction techniques as illustrated in figure \ref{fig:CT-RATE}.

The authors also introduced the CT-CLIP framework, which leverages contrastive language-image pre-training to enable broad applicability without requiring task-specific fine-tuning. This framework excels in multi-abnormality detection and case retrieval tasks, demonstrating robust performance not only on data with similar distributions but also under distribution shifts.

By integrating CT-CLIP's vision encoder with a pretrained large language model (LLM), the authors developed CT-CHAT, a model tailored for 3D chest CT volumes. CT-CHAT outperforms other multimodal AI assistants, such as LLaVa 1.6, LLaVa-Med, and CXR-LLaVa, in generating both concise and detailed responses, handling multiple-choice questions, and producing high-quality radiology reports. The strength of CT-CHAT lies in its ability to utilize 3D spatial information processed by CT-CLIP’s pretrained vision transformer, which captures more intricate anatomical details compared to 2D models. Additionally, the integration of 3D CT volumes with radiology-specific linguistic data from CT-RATE enables CT-CHAT to deliver highly accurate and clinically relevant outputs, surpassing models trained on general-purpose text data.

\subsection{OmniMedVQA}
OmniMedVQA\cite{hu2024omnimedvqanewlargescalecomprehensive} is a large-scale evaluation benchmark designed for medical Visual Question Answering. It includes 118,010 images and 127,995 question-answer items, covering different modalities across more than anatomical regions, sourced from 73 distinct medical datasets. This dataset aims to facilitate the evaluation of Large Vision-Language Models (LVLMs) in medical applications.  

The benchmark encompasses various imaging modalities, including: Magnetic Resonance Imaging (MRI), Computed Tomography (CT), X-Ray, histopathology, fundus photography, digital photography, ultrasound, endoscopy, dermoscopy, Optical Coherence Tomography (OCT), Infrared Reflectance Imaging (IRI) and colposcopy.

\subsection{Evaluation metrics}
\begin{table*}[ht]
    \centering
    \caption{\textbf{Evaluation metrics of Medical VLMs}}
    \begin{tabular}{lll}%{p{1.6cm}p{3cm}p{3cm}}
        \toprule
        \textbf{Evaluation Metrics} & \textbf{Description} & \textbf{Use Case} \\
        \midrule
        BLEU & N-gram overlap for text similarity. & General VQA evaluation. \\
        % \hline
        ROUGE & Recall-based keyword overlap. & Long medical descriptions. \\
        % \hline
        BERTScore & Semantic similarity using BERT embeddings. & Semantic relevance in medical text. \\
        % \hline
        CheXpert Labeler & Rule-based pathology detection. & Chest X-ray report evaluation. \\
        % \hline
        RadGraph & Graph-based entity-relationship evaluation. & Radiology report accuracy. \\
        % \hline
        Clinical Correctness Score & Expert evaluation of diagnostic accuracy. & Clinical report validation. \\
        \bottomrule
    \end{tabular}
    \label{tab:metrics}
\end{table*}
\subsubsection{Medical VQA metrics}
The performance of Visual Language Models (VLMs) on Visual Question Answering (VQA) tasks in the medical domain is evaluated based on how accurately and coherently they generate textual interpretations of medical images in response to specific questions. These metrics assess various aspects of the generated text in relation to a reference answer, measuring factors such as lexical similarity, semantic relevance, and clinical accuracy.

\paragraph{BLEU Score} BLEU measures the n-gram overlap between the model's generated response and a reference answer. While commonly used in general NLP tasks, in medical VQA, it highlights the degree of exact phrase matching, though it may not fully capture semantic relevance in complex medical descriptions.

Although this metric is well-established in NLP and effective for comparing concise, factual responses, they have limitations in capturing clinical significance. As shown by \cite{wieting_etal_2019_beyond}, it fails to assign partial credit for semantically correct answers that differ lexically from the reference, have limited output ranges, and can penalize clinically accurate responses if phrasing differs. This is especially pertinent in medical VQA, where correct clinical information is crucial, even if expressed differently from the reference text.

\paragraph{ROUGE Score}: ROUGE evaluates recall by measuring the overlap of key words or phrases, emphasizing the capture of critical content in the response. For medical VQA, it ensures the model includes essential terms that may be crucial for interpreting medical findings. Contrary to BLEU, ROUGE is better for longer medical descriptions and capture content coverage. But there are still questions about its relevance; since it measures overlap between generated and reference texts based on n-grams, it may fail to capture the nuanced, specialized language of medical contexts. Additionally, it can overlook clinical accuracy, focusing on surface-level similarities rather than the semantic or clinical relevance of terms, which can lead to misleadingly high scores even when hallucinations or inaccuracies are present in generated reports \cite{jiang2024comtchainofmedicalthoughtreduceshallucination}.

\paragraph{BERTScore} This metric leverages pre-trained BERT embeddings to compare the semantic similarity between the generated response and the reference answer. BERTScore is beneficial in medical contexts because it evaluates meaning beyond surface-level text matching, helping to assess the model's understanding of complex medical language. BERTScore can struggle with medical VLMs because it primarily focuses on semantic similarity, often missing finer distinctions crucial for medical accuracy. It may also underperform in scenarios where precise domain knowledge is necessary to assess the relevance of generated content, leading to challenges in accurately evaluating models used for clinical report generation or interpretation of radiological images  \cite{jiang2024comtchainofmedicalthoughtreduceshallucination}.

\paragraph{Clinical Accuracy Score} This metric assesses the clinical correctness of the generated response, going beyond textual similarity. It evaluates whether the response aligns with accurate medical knowledge and terminology, which is essential for medical applications where the precision of information is critical.

\subsubsection{Domain-Specific Metrics}
\paragraph{CheXpert Labeler}The CheXpert Labeler\cite{irvin2019} is a rule-based system designed to evaluate the presence or absence of various pathologies in chest X-ray reports. It automates the annotation process by identifying clinical findings such as pneumothorax, edema, and cardiomegaly. For report generation, this metric can serve as a validation tool by assessing whether generated reports accurately include or omit relevant clinical observations. The CheXpert Labeler has become widely used due to its reliability in automated evaluation for large-scale datasets, such as MIMIC-CXR\cite{Johnson2019}, enabling systematic comparison between generated and reference reports in clinical language models.

\paragraph{RadGraph} RadGraph\cite{jain2021radgraphextractingclinicalentities} is a graph-based evaluation framework designed to capture anatomical and pathological relationships in radiology reports. It structures reports into a directed graph of entities and relationships, representing anatomical locations, conditions, and their attributes. This enables more precise assessment of generated reports, especially for clinical accuracy and relational correctness. RadGraph is useful for models that need to convey not only individual findings but also their relationships, such as locating pathologies in specific anatomical regions, which is crucial in radiology.

\paragraph{Clinical Correctness Score} The Clinical Correctness Score (CCS)\cite{zhang2020radiologyreportgenerationmeets} involves expert human evaluators assessing the diagnostic accuracy and appropriateness of generated reports. This score addresses potential hallucinations and ensures that generated content aligns with medical standards, focusing on whether diagnoses, observations, and recommendations reflect clinical expertise. CCS is invaluable for assessing the real-world applicability of generated reports, as it captures nuances that automated metrics might miss.

\paragraph{Human evaluation metrics}
Human evaluation metrics are critical in the medical domain, especially for tasks such as report generation, where clinical accuracy and contextual relevance cannot be fully captured by automated metrics alone. Key human evaluation metrics commonly used for assessing the quality of medical language models are: (i) Content Quality and Relevance to determine whether the generated report includes the necessary medical findings, diagnoses, and insights relevant to the clinical task \cite{Johnson2019, zhu2024leveragingprofessionalradiologistsexpertise}; (ii) Fluency and Naturalness referring to the grammatical correctness, readability, and flow of the generated report, ensuring it resembles a report written by a clinical professional\cite{qin_song_2022_reinforced}; (iii) Diagnostic Accuracy; (iv) Coherence with Image; (v) Comprehensiveness and Coverage; (vi) etc.

% https://arxiv.org/abs/2312.07867

\section{Challenges and Limitations}
The rapid advancement of VLMs in medicine has demonstrated their potential to revolutionize tasks such as medical image analysis, report generation, and visual question answering. However, significant challenges remain that hinder their widespread adoption and effectiveness in clinical settings. These limitations stem from issues such as data scarcity, narrow task generalization, lack of interpretability, ethical concerns, computational resources and the difficulty of integrating VLMs into clinical workflows.

\subsection{Scarcity and quality of medical datasets}
 A critical limitation remains the scarcity of high-quality, task-specific datasets, particularly for specialized applications like visual question answering (VQA) and report generation. For instance, while large-scale datasets such as Medtrinity-25M (25M samples across radiology, pathology, and EHRs) and PMC-VQA (1.5M VQA pairs) demonstrate progress in data volume, many widely used benchmarks—including VQA-RAD (3k samples), SLAKE (2k samples), and PathVQA (6k samples)—remain limited in size and diversity\cite{Lau2018,liu2021slakesemanticallylabeledknowledgeenhanceddataset,he2020pathvqa30000questionsmedical}. These smaller datasets often lack representation of rare conditions or underrepresented modalities, leading to biases in model performance. For example, CheXpert and MIMIC-CXR-JPG focus predominantly on chest X-rays, limiting their utility for training models to interpret MRI or CT scans (Irvin et al., 2019; Johnson et al., 2019). This imbalance is further compounded by the labor-intensive process of medical data annotation, which relies heavily on expert clinicians and radiologists, as seen in datasets like CheXpert Plus\cite{chambon2024chexpertplusaugmentinglarge}.

 \subsection{Narrow scope of task generalization}
 While benchmarks such as GMAI-MMBench (26k samples across 38 modalities) and OmniMedVQA (128k samples spanning 12 modalities) aim to evaluate multi-modal understanding, most models are still trained and validated on modality-specific tasks \cite{chen2024gmaimmbenchcomprehensivemultimodalevaluation,hu2024omnimedvqanewlargescalecomprehensive}. For instance, RadBench (137k samples across 6 modalities) emphasizes image-text alignment but does not fully address the complexity of real-world clinical workflows, where models must integrate heterogeneous data types like EHRs, pathology reports, and imaging studies \cite{kuo2024radbenchevaluatinglargelanguage}. This specialization limits the adaptability of VLMs to cross-modal tasks, such as correlating radiology findings with pathology results, a gap partially addressed by Medtrinity-25M but still underexplored in practice \cite{xie2024medtrinity25mlargescalemultimodaldataset}.

\subsection{Lack of interpretability and trust}
Despite advancements in models like PMC-VQA and Med-Flamingo, which generate detailed answers to medical questions, the reasoning processes behind these outputs are often opaque \cite{Zhang2024, moor2023medflamingomultimodalmedicalfewshot}. Benchmarks such as BESTMVQA and CT-RATE evaluate diagnostic accuracy but lack metrics to assess the clinical plausibility or explainability of model predictions, leaving clinicians skeptical of AI-generated reports \cite{hong2023bestmvqabenchmarkevaluationmedical,hamamci2024developinggeneralistfoundationmodels}. Furthermore, ethical and privacy concerns persist, particularly with datasets sourced from EHRs (e.g., Medtrinity-25M) or hospital repositories (e.g., GMAI-MMBench), where de-identification protocols may not fully eliminate re-identification risks \cite{xie2024medtrinity25mlargescalemultimodaldataset,chen2024gmaimmbenchcomprehensivemultimodalevaluation}.

Many of these models operate as "black boxes," making it difficult for clinicians to understand how they arrive at specific diagnoses or recommendations. For example, while models like MedViLL and BioViL demonstrate strong performance in tasks such as medical image captioning and visual question answering, their decision-making processes remain opaque \cite{Moon_2022, Boecking2022}. This lack of transparency can hinder clinician trust and adoption, particularly in high-stakes medical applications where understanding the rationale behind a diagnosis is critical. Efforts to improve interpretability, such as attention visualization or explainable AI (XAI) techniques, are still in their infancy and require further development to bridge this gap\cite{radford2021learningtransferablevisualmodels, Zhang2024}

\subsection{Ethical and privacy concerns}
The use of patient data for training these models raises issues related to data privacy and compliance with regulations such as HIPAA. For instance, datasets like MIMIC-CXR, while de-identified, still contain sensitive patient information, and their use in training VLMs must be carefully managed to avoid privacy breaches \cite{Johnson2019}. Additionally, there is a risk of perpetuating biases present in the training data, which could lead to disparities in healthcare outcomes for different demographic groups. For example, if a VLM is trained predominantly on data from a specific population, it may underperform when applied to patients from underrepresented groups, exacerbating existing healthcare inequities \cite{Boecking2022,Zhang2024}.

\subsection{Computational resources required}
The computational resources required to train and deploy VLMs are another limitation. Training state-of-the-art models like BLIP-2, LLaVA, and MiniGPT-4 requires substantial computational power, often involving large-scale GPU clusters and extensive training times \cite{li2023blip2bootstrappinglanguageimagepretraining,liu2023visualinstructiontuning,zhu2023minigpt4enhancingvisionlanguageunderstanding}. This makes it challenging for resource-constrained healthcare institutions to adopt these technologies. Furthermore, the need for continuous fine-tuning and updates to keep the models relevant adds to the computational burden, limiting their scalability and accessibility in low-resource settings\cite{Zhang2024,luo2024buildinguniversalfoundationmodels}.

\subsection{Difficulty of integrating VLMs into clinical workflows}
While the developed models show promise in tasks such as medical image analysis and report generation, their real-world application requires seamless integration into existing electronic health record (EHR) systems and clinical workflows. For example, models like Flamingo-CXR and Med-Flamingo have demonstrated the ability to generate radiology reports, but their adoption in clinical practice depends on their ability to provide real-time, actionable insights without disrupting existing workflows \cite{Tanno2024,moor2023medflamingomultimodalmedicalfewshot}. Clinicians may also require additional training to use these tools effectively, further complicating their integration into routine practice.

\section{Opportunities and Future Directions}
Despite the challenges, the field of medical VLMs is ripe with opportunities for innovation. By leveraging large-scale multimodal datasets, improving cross-modal generalization, and prioritizing interpretability and ethical considerations, researchers can develop models that are not only more accurate but also more aligned with clinical needs. These advancements have the potential to transform medical workflows, enhance diagnostic accuracy, and improve patient outcomes, paving the way for a new era of AI-driven healthcare.

\subsection{Scaling and diversifying datasets}
 Scaling and diversifying datasets could address current limitations in data quality and task specificity. For example, Medtrinity-25M’s integration of radiology, pathology, and EHR data provides a template for building comprehensive multimodal datasets that bridge imaging, text, and clinical context \cite{xie2024medtrinity25mlargescalemultimodaldataset}. Similarly, synthetic data generation techniques, as proposed in VividMed, could augment smaller benchmarks like VQA-RAD or SLAKE, enhancing their utility for training robust models \cite{luo2024vividmedvisionlanguagemodel}. Collaborative efforts to expand datasets such as PMC-VQA (1.5M samples) and OmniMedVQA (128k samples) could also improve coverage of rare diseases and underrepresented modalities \cite{Zhang2024,hu2024omnimedvqanewlargescalecomprehensive}.

\subsection{Cross-modal generalization}
Benchmarks like GMAI-MMBench, which evaluate 18 tasks across 38 modalities, highlight the need for models capable of integrating diverse data types \cite{chen2024gmaimmbenchcomprehensivemultimodalevaluation}. Techniques such as contrastive learning, employed in ConVIRT and MedCLIP, could be extended to align features from radiology, pathology, and EHRs, enabling VLMs to perform tasks like correlating imaging findings with lab results \cite{zhang2022contrastivelearningmedicalvisual,wang2020vdbertunifiedvisiondialog}. Additionally, frameworks like LLaMA-Adapter-V2, which use parameter-efficient tuning to adapt models to new modalities, could facilitate rapid deployment across clinical settings without requiring retraining from scratch \cite{gao2023llamaadapterv2parameterefficientvisual}.

\subsection{Interpretability and clinical relevance}
Advancing interpretability and clinical relevance will require rethinking evaluation metrics. For instance, benchmarks such as RadBench and BESTMVQA could incorporate criteria for explainability, such as attention map accuracy or alignment with clinical guidelines \cite{kuo2024radbenchevaluatinglargelanguage,hong2023bestmvqabenchmarkevaluationmedical}. Models like VividMed, which localize anatomical regions during diagnosis, demonstrate how visual grounding mechanisms can make outputs more transparent \cite{luo2024vividmedvisionlanguagemodel}. Similarly, integrating expert feedback loops, as seen in Flamingo-CXR’s human evaluation framework, could refine model outputs to better match clinician expectations \cite{Tanno2024}.

\subsection{Ethical and computational opportunities}
Finally, addressing ethical and computational challenges will be essential for real-world deployment. Federated learning approaches, as explored in BiomedGPT, could enable training on distributed datasets like Medtrinity-25M without compromising patient privacy \cite{Zhang2024}. Lightweight architectures such as DeepSeek-VL (1.3B parameters) and LLaMA-Adapter-V2, which reduce computational costs through token compression and parameter-efficient tuning, offer pathways to democratize access to VLMs in resource-constrained settings \cite{lu2024deepseekvlrealworldvisionlanguageunderstanding,gao2023llamaadapterv2parameterefficientvisual}. By aligning technical advancements with clinical needs—such as seamless EHR integration and real-time decision support—future VLMs could transform medical workflows while adhering to ethical and regulatory standards.

\section{Conclusion}
Medical Vision-Language Models (VLMs) represent a transformative advancement in AI-driven healthcare, offering the potential to enhance medical image interpretation, automated reporting, and clinical decision-making. However, several challenges hinder their widespread adoption and effectiveness. These include limitations in semantic understanding, dataset biases, computational constraints, and the inadequacy of traditional evaluation metrics in assessing clinical relevance. Addressing these issues is critical to ensuring that VLMs provide reliable, accurate, and contextually meaningful insights in real-world medical applications.

Despite these challenges, the future of medical VLMs is promising. Advances in pretraining methodologies, multimodal integration, and robust benchmarking will enhance model performance and generalizability. The incorporation of diverse data sources, such as electronic health records and genomic data, will further improve the contextual understanding of patient health. Additionally, mitigating dataset biases through diverse and representative training data will enable VLMs to perform effectively across varied patient populations and imaging modalities.

Furthermore, the seamless integration of VLMs into clinical workflows through intuitive and physician-friendly interfaces will foster trust and adoption among healthcare professionals. Expanding applications in digital pathology, AI-driven image retrieval, and precision medicine will unlock new opportunities for improving patient outcomes. Ultimately, by addressing current limitations and capitalizing on emerging advancements, medical VLMs can revolutionize healthcare, making AI-powered medical imaging and diagnosis more accurate, accessible, and impactful.

\section*{Acknowledgment}
The authors would like to thank the Fatima Fellowship and its founders for providing the opportunity and platform for this research collaboration.

\ifCLASSOPTIONcaptionsoff
  \newpage
\fi

\newpage
\bibliographystyle{IEEEtran.bst}
\bibliography{references}

% \end{IEEEbiography}


% % insert where needed to balance the two columns on the last page with
% %\newpage



%\vfill


\end{document}